\documentclass[10pt,twocolumn,letterpaper]{article}

\usepackage{cvpr}              

\usepackage{amsmath}
\usepackage{amssymb}
\usepackage[ruled,vlined]{algorithm2e}
\usepackage[most]{tcolorbox}
\usepackage{graphicx}
\usepackage{caption}
\usepackage{multirow}
\usepackage{booktabs}
\usepackage[table]{xcolor} 
\definecolor{lightyellow}{RGB}{255, 255, 204}
\definecolor{lightpink}{RGB}{255, 230, 234}

\definecolor{cvprblue}{rgb}{0.21,0.49,0.74}
\usepackage[pagebackref,breaklinks,colorlinks,allcolors=cvprblue]{hyperref}

\def\paperID{*****} 
\def\confName{CVPR}
\def\confYear{2026}

\title{RAZOR: Ratio-Aware Layer Editing for Targeted Unlearning in Vision Transformers and Diffusion Models}

\author{
Ravi Ranjan\thanks{Corresponding author \\Accepted to the \textbf{CVPR 2026}. \\ To appear in the Findings Track Proceedings of IEEE/CVF Conference. \\  Code available at \url{https://github.com/raviranjan-ai/RAZOR-cvpr2026}.}\\
Florida International University\\
Miami, USA\\
{\tt rkuma031@fiu.edu}
\and
Utkarsh Grover\\
University of South Florida\\
Tampa, USA \\
{\tt\small utkarshgrover@usf.edu}
\and
Xiaomin Lin\\
University of South Florida\\
Tampa, USA \\
{\tt\small xlin2@usf.edu}
\and
Agoritsa Polyzou\\
Florida International University\\
Miami, USA\\
{\tt\small apolyzou@fiu.edu}
}

\begin{document}
\maketitle

\begin{abstract}
Transformer-based diffusion and vision-language models have achieved remarkable success; yet, efficiently removing undesirable or sensitive information without retraining remains a central challenge for model safety and compliance. We introduce \textbf{R}atio-\textbf{A}ware \textbf{Z}ero/One-step \textbf{O}ptimized \textbf{R}etentive unlearning (\textbf{RAZOR}), a lightweight, model-agnostic unlearning framework that generalizes forgetting updates to coordinated multi-layer and multi-head edits within transformer backbones. 
RAZOR identifies the most important layers and attention heads by measuring how much they contribute to forgetting the target data while preserving useful knowledge. Then, it updates these parts of the model using a carefully regularized rule to avoid harming overall performance. The set of edited components grows gradually, ensuring precise unlearning without over-editing or damaging unrelated capabilities.
We evaluate RAZOR on CLIP, Stable Diffusion, and vision-language models (VLMs) using widely adopted unlearning benchmarks covering identity, style, and object erasure tasks. Our results show that RAZOR achieves highly accurate and stable forgetting, even under quantization. 
This approach offers stronger retention and better efficiency than prior methods. Notably, it also operates significant faster than conventional techniques. These results demonstrate that RAZOR is a practical and scalable solution for safe, adaptive unlearning in transformer-based vision models.
\end{abstract}

\begin{figure}[t]
   \includegraphics[width=\linewidth]{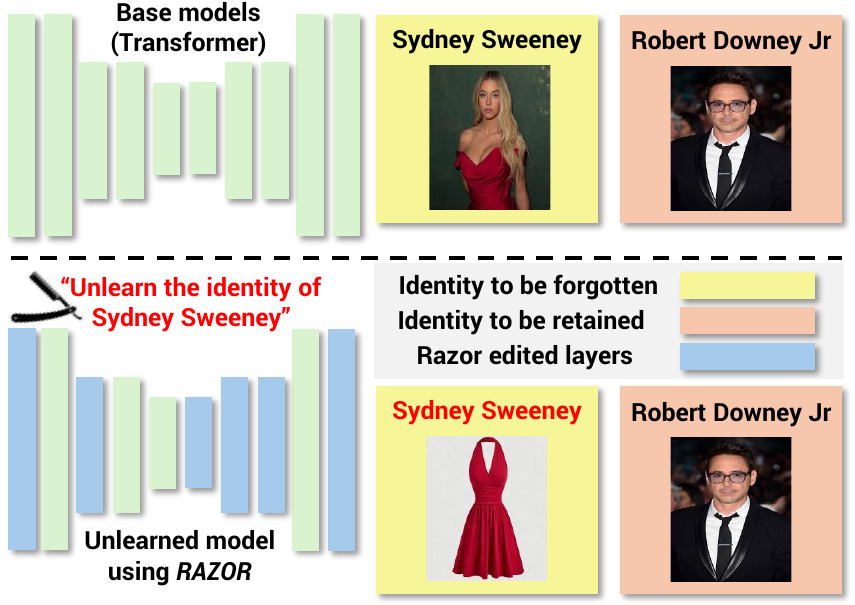}
   \caption{RAZOR edits a subset of layers to unlearn selected visual identity while retaining others. And RAZOR operates in a model agnostic way across CLIP, vision language models(VLMs), and diffusion models.}
   \label{fig:intro}
   \vspace{-0.6cm}
\end{figure}

\section{Introduction}
\label{sec:intro}
\vspace{-0.2cm}
Modern large-scale vision language and generative models from contrastive encoders, like CLIP~\cite{radford2021clip}, to text-to-image diffusion models, like Stable Diffusion~\cite{rombach2022high}, and multi-modal assistants, such as LLaVA~\cite{liu2024improved}, are trained on massive datasets. While this yields powerful capabilities, it also embeds sensitive or unwanted information, raising serious privacy and compliance concerns (e.g., the GDPR~\cite{zarsky2016incompatible}). When certain data must be removed from a model (e.g., due to user requests or legal mandates), retraining from scratch on a filtered dataset is prohibitively expensive~\cite{cao2015towards}. \textit{Machine unlearning} has emerged as a compelling alternative~\cite{nguyen2022survey}, referring to techniques for \textit{post hoc} removal of targeted knowledge from a trained model without full retraining.

The figure \ref{fig:intro} shows how RAZOR selectively edits only a small set of layers to erase a specific identity while keeping other visual identities intact across transformer-based vision models.
An ideal unlearning method should satisfy three key criteria: (a) \textbf{efficiency}: removing targeted knowledge with minimal computational cost, (b) \textbf{precision}: erasing only the intended information without affecting unrelated content, and (c) \textbf{robustness}: ensuring the forgotten data cannot be recovered, even under adversarial or shifted inputs. These properties are essential for building trustworthy and scalable unlearning systems ~\cite{thudi2022unrolling, fan2024salun}.

Achieving all three is challenging, and existing approaches typically sacrifice one aspect for another~\cite{fan2024salun, foster2024loss}. For example, conventional fine-tuning-based unlearning or direct gradient ascent on the forget set~\cite{thudi2022unrolling} can forget the target but often at the cost of severe utility drop. Recent selective update strategies, such as Saliency Unlearning (SalUn)~\cite{fan2024salun} and Selective Synaptic Dampening (SSD)~\cite{foster2024fast}, improve efficiency by adjusting only high-saliency weights. However, these still require expensive multi-layer updates and can degrade unrelated concepts. A complementary direction confines changes to a single layer. \textsc{SLUG}~\cite{cai2025slug} pioneered this idea by editing one critical layer per request, dramatically reducing compute. Yet such \textit{single-layer edits} are brittle: if knowledge is distributed, editing one layer may be insufficient~\cite{thudi2022unrolling}.

To address these limitations, we introduce \textbf{RAZOR} (Ratio-Aware Zero/One-step Optimized Retentive unlearning), a general, ratio-aware multi-layer editing framework for targeted unlearning in transformer-based vision and diffusion models. RAZOR generalizes one-shot unlearning to coordinated \textbf{multi-layer} and multi-head updates while explicitly controlling the trade-off between forgetting and retaining. We evaluate RAZOR across a broad suite of unlearning tasks and model families, including identity and object removal in CLIP, as well as concept erasure in Stable Diffusion and vision–language models (VLMs). We additionally apply model quantization to assess robustness, following the observations of Zhang et al.~\cite{zhang2024catastrophic}, who highlighted the risk of quantization-induced degradation in unlearning performance.
By unifying principled gradient scoring with efficient multi-layer editing, RAZOR provides a practical and scalable framework for safe, precise, and compute-efficient unlearning in large transformer-based systems. 
It delivers state-of-the-art performance in complete and quantization stable forgetting, while simultaneously preserving higher retention accuracy without additional computational cost.
The key contributions of RAZOR are threefold:
\begin{enumerate}
    \item \textbf{ Ratio-aware gradient scoring}, which computes one-shot forget and retain gradients per layer or attention head and ranks them using a \textbf{forget retain significance ratio}. This ratio allows RAZOR to identify components that exert high forgetting influence while minimally disrupting retained knowledge, enabling more principled selection of edits compared to single-layer approaches.
    \item \textbf{Constrained multi-objective loss}, wherein we design a unified unlearning objective that jointly optimizes for target forgetting, retention preservation, and mismatch regularization, ensuring both effectiveness and stability. 
    \item \textbf{Iterative refinement}, that dynamically expands and updates the edited parameter set until the desired forgetting threshold is achieved, avoiding over-editing and limiting collateral degradation.
\end{enumerate}

\section{Related Work}
\label{sec:related-work}

Early unlearning methods for vision-language systems relied on second order optimization. Selective Synaptic Dampening (SSD) \cite{foster2024fast} used Fisher information to identify and attenuate forget-critical parameters, achieving strong erasure but at prohibitive computational cost and with notable utility loss on complex datasets. LoTUS \cite{spartalis2025lotus} improved efficiency by shifting to output space smoothing, regularizing predictions toward uniform distributions and avoiding curvature computation. However, this shortcut induced embedding drift, degrading retrieval consistency a critical issue for representation aligned applications. SalUn \cite{fan2024salun} addressed both inefficiency and instability through gradient based parameter saliency, updating only weights most influenced by the forget set. This maintained high utility and stable embeddings but produced incomplete forgetting due to overly conservative selection: parameters were ranked using only forget gradients, leaving residual memorization.
Extending these methods beyond discriminative architectures is inherently difficult, as generative systems distribute semantic knowledge across layers, embeddings, and attention maps, making localized parameter edits ineffective.

SLUG \cite{cai2025slug} prioritized efficiency by restricting edits to a single high-impact layer, preserving utility and stability but failing when target knowledge was distributed across multiple components. In such cases, localized editing proved too brittle for com-positionally encoded concepts. Erased Stable Diffusion (ESD) \cite{gandikota2023erasing} introduced negative guidance during sampling to suppress concepts, though without true removal. Later methods Forget Me Not (FMN) \cite{zhang2024forget}, Unified Concept Editing (UCE) \cite{gandikota2024unified}, and EDiff \cite{xie2025ec} incorporated embedding interventions and dynamic masking to balance unlearning completeness and content fidelity, yet the trade off persisted.

Across these methods, a common limitation emerges: parameter selection is driven solely by forget set saliency, while retention conflicts are mitigated only afterward. This sequential strategy cannot avoid coupling between forget and retain dynamics. \textbf{RAZOR} departs from this paradigm with ratio-aware scoring that jointly evaluates forget pressure and retain alignment during parameter selection, enabling multi-layer edits with explicit trade-off control across both vision-language and diffusion architectures. For detailed related work refer Appendix \ref{app:related}. 

\section{RAZOR: Ratio-Aware Layer/Head Unlearning}
\label{sec:method}

\begin{figure*}[tb]
    \centering
    \includegraphics[width=0.95\linewidth]{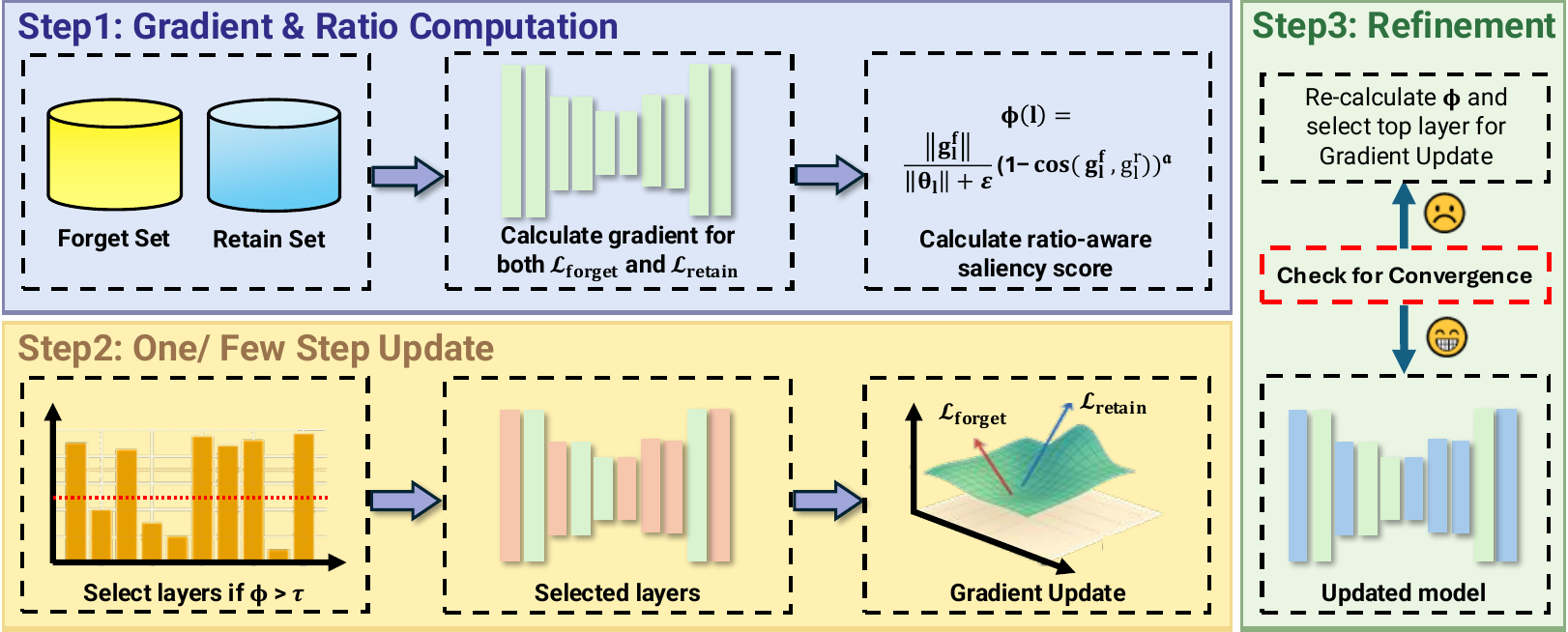} 
    \caption{Overview of the RAZOR unlearning pipeline across gradient computation, selective updates, and refinement.}
    \label{fig:Main}
    \vspace{-0.5cm}
\end{figure*}

\textbf{RAZOR} is a general-purpose framework for efficient machine unlearning that targets multiple influential layers or attention heads. It selectively updates these components along carefully composed gradient directions, guided by an explicit \emph{forget/retain ratio}. In contrast to single-layer methods such as SLUG~\cite{cai2025slug} or prior saliency- and damping-based strategies~\cite{fan2024salun,foster2024fast}, RAZOR: (i) scores components using a ratio-aware saliency that accounts for forgetting pressure and retention alignment; (ii) disentangles learning objectives through a three-part loss that explicitly controls utility, forgetting, and stability; and (iii) applies uniformly across vision encoders (ViT), text encoders (CLIP), diffusion-based encoders, and vision-language models. The figure~\ref{fig:Main} illustrates how RAZOR identifies high-impact layers using ratio-aware saliency, applies targeted gradient updates, and iteratively refines the model until convergence.

\subsection{Preliminaries}
Let $f_\theta$ denote a pretrained model with parameters $\theta$ and indexed layers/heads $\mathcal{L} = \{1, \dots, L\}$. Given a \emph{forget set} $\mathcal{D}_f$ and a \emph{retain set} $\mathcal{D}_r$ with respective sizes $N_f$ and $N_r$, the objective is to produce updated parameters $\theta^\star$ such that information specific to $\mathcal{D}_f$ is removed while performance on $\mathcal{D}_r$ and unrelated data is preserved.

For contrastive vision-language models (e.g., CLIP~\cite{radford2021clip}) and for an image-text pair $(x_i, y_i)$, there is a text encoder and an image encoder which generate the corresponding normalized image/text embeddings $(v_i, t_i)$. In diffusion models~\cite{rombach2022high}, the text encoder outputs $e = f_t(t_i)$ to condition the UNet. For vision-language models (e.g., LLaVA~\cite{liu2024improved}), the visual encoder produces tokens consumed by a downstream LLM.

\subsection{Ratio-Aware Layer/Head Selection}
\label{sec:ratioaware}
The first step of our method is to identify salient components. To do so, we compute one-shot gradients for each $l \in \mathcal{L}$:
\begin{equation}
g^f_l = \nabla_{\theta_l} \mathcal{L}_{\text{forget}},
\qquad
g^r_l = \nabla_{\theta_l} \mathcal{L}_{\text{retain}}.
\end{equation}
We define a \textbf{ratio-aware saliency score}:
\begin{equation}
\label{eq:ratio_score}
\phi(l)
=
\frac{\|g^f_l\|_2}{\|\theta_l\|_2+\varepsilon}
\cdot
\big(1-\cos(g^f_l,\,g^r_l)\big)^\alpha,
\end{equation}
with hyper-parameters $\alpha \in [0,1]$ (trade-off between magnitude and orthogonality) and $\varepsilon > 0$ (numerical stability). A high $\phi(l)$ indicates strong forgetting influence and low retention alignment, suggesting low collateral damage. We select the subset of layers whose ratio-aware saliency exceeds the threshold $\tau$, forming  
$\mathcal{K} = \{\, l \in \mathcal{L} \mid \phi(l) > \tau \,\}$,  
where $\tau$ controls the minimum significance required for a layer to participate in RAZOR updates.

\subsection{Three-Loss Objective with Ratio Control}
We minimize a composite loss that explicitly balances forgetting and retention, controlled by a ratio hyperparameter $\rho \in (0,1]$:
\begin{align}
\label{eq:razor_obj}
\min_{\theta} \;\; 
\mathcal{L}_{\text{RAZOR}}(\theta)
:= &\;
\underbrace{\mathcal{L}_{\text{retain}}(\theta;\mathcal{D}_r)}_{\text{utility } \uparrow}
\;+\;
\lambda_f \rho \,
\underbrace{\mathcal{L}_{\text{forget}}(\theta;\mathcal{D}_f)}_{\text{forgetting } \uparrow \text{ (via ascent)}} \nonumber\\
&\;+\;
\lambda_m \,
\underbrace{\mathcal{L}_{\text{mismatch}}(\theta;\mathcal{D}_f,\mathcal{D}_r)}_{\text{stability}},
\end{align}
where $\lambda_f, \lambda_m > 0$ control forgetting pressure and regularization, respectively. Forgetting is implemented by ascending $\nabla_\theta \mathcal{L}_{\text{forget}}$, or equivalently minimizing $-\mathcal{L}_{\text{forget}}$. 

Our composite function is modular and model-agnostic. The concrete form of each loss term depends on the type of representation provided by the underlying model used (i.e., vision–language embeddings in CLIP, diffusion latents in Stable Diffusion, or autoregressive logits in VLMs). Next, we \textit{present the losses related to CLIP}, while we present the functions used for the rest of the underlying models in Sect.~\ref{sec:method:instantiation}.

\textbf{(a) Retain loss $\mathcal{L}_{\text{retain}}$.} ensures controlled, localized edits that only change targeted information, without affecting the rest of the model. To preserve utility, we optimize task-relevant alignment on the retain dataset, $\mathcal{D}_r$. 
For transformers and the CLIP model, we adopt the symmetric InfoNCE loss:

\begin{align}
\label{eq:retain}
\mathcal{L}_{\text{retain}}
&=
\frac{1}{2N_r}\sum_{i=1}^{N_r}
\Bigg[
-\log \frac{\exp(\langle v_i,t_i\rangle/\tau)}{\sum_{j=1}^{N_r}\exp(\langle v_i,t_j\rangle/\tau)}
\Bigg. \nonumber\\
&\hspace{6em}
\Bigg.
-\log \frac{\exp(\langle v_i, t_i\rangle/\tau)}{\sum_{j=1}^{N_r}\exp(\langle v_j, t_i\rangle/\tau)}
\Bigg],
\end{align}

where $\tau > 0$ is the temperature~\cite{radford2021clip}. The two terms inside the parentheses capture if the image embedding is close to the right text, and at the same time, if the text embedding should also be close to the right image.

\textbf{(b) Forget loss $\mathcal{L}_{\text{forget}}$.}
To erase knowledge of the forget dataset, $\mathcal{D}_f$, we drive apart aligned pairs $(v_i, t_i)$ using cosine similarity embedding loss:
\begin{equation}
\label{eq:forget}
\mathcal{L}_{\text{forget}}
=
\frac{1}{N_f}\sum_{i=1}^{N_f}
\big(1-\langle v_i,t_i\rangle\big),
\end{equation}
which reduces similarity between positive image-text pairs~\cite{cai2025slug}. 

\textbf{(c) Mismatch loss $\mathcal{L}_{\text{mismatch}}$.} It pushes the model to change its embeddings (and their similarity) for the image and text pairs we need to forget.
We regularize changes in similarity from the initial, frozen model:
\begin{equation}
\label{eq:mismatch}
\mathcal{L}_{\text{mismatch}}
=
\frac{1}{N_f}\sum_{i=1}^{N_f}
\left(\langle v_i,t_i\rangle - \langle v_i^{(0)}, t_i^{(0)}\rangle
\right),
\end{equation}
where $v_i^{(0)}, t_i^{(0)}$ are the embeddings of the initial model, before any editing for unlearning. For stable diffusion and VLMs, those correspond to forget-set negatives and logit margins, respectively. This controls latent drift and improves both privacy. Details provided in table~\ref{tab:meth-1}.

\subsection{One-/Few-Step Updates}
\label{sec:method:updates}
For each selected $l \in \mathcal{K}$, we apply a blended gradient update:
\begin{equation}
\label{eq:update}
\Delta\theta_l
=
-\eta_l
\left(
-\lambda_f\rho\, g^f_l
+
g^r_l
+
\lambda_m\nabla_{\theta_l}\mathcal{L}_{\text{mismatch}}
\right),
\end{equation}

Here, $\eta_l$ denotes the effective per-layer learning rate that scales the blended gradient update, while $\lambda_l \in \mathbb{R}_{+}$ parameterizes this step magnitude for layer $l$. We determine $\lambda_l$ via a lightweight binary search that selects the largest stable step size yielding the best forgetting retention trade-off on a small validation subset.
This preserves efficiency~\cite{cai2025slug} while generalizing to multi-layer/head edits.

\subsection{Iterative Growing of \texorpdfstring{$\mathcal{K}$}{K}}
\label{sec:iter}
If after updating the initial selection of layers, $\mathcal{K}$, the resulting model $\theta'$ 
does not satisfy the threshold criterion, RAZOR performs an iterative refinement process that adaptively grows $\mathcal{K}$ to add more layers as needed, $\mathcal{K} \leftarrow \mathcal{K} \cup \{l\}$. Our goal is to update additional layers until the model reaches a desirable unlearning performance (or a fixed iteration limit is reached, $t\leq6$).
At each iteration $t$, we:
\begin{enumerate}
    \item Recompute the saliency scores $\phi_t(l)$ using the updated parameters $\theta^{(t)}$.
    \item Grow the active set by adding one new layer $l$ at a time that has the highest $\phi_t(l)$.
    \item Update the model parameters for the selected layer $l$, like in Sect.~\ref{sec:method:updates}.
\end{enumerate}

In practice, this ensures stable and progressive identification of the most influential layers for unlearning while avoiding premature convergence.

\subsection{Instantiation Across Model Families}
\label{sec:method:instantiation}

\begin{table}[bt]
\centering
\renewcommand{\arraystretch}{1.5}

\small
\caption{RAZOR loss components across model families. Overall, $\mathcal{L}_{\text{RAZOR}} = \mathcal{L}_{\text{retain}} + \lambda_f \rho\,\mathcal{L}_{\text{forget}} + \lambda_m \mathcal{L}_{\text{mismatch}}$ 
uses standard training objectives and unlearning losses from prior work. SDR = Similarity Drift Regularizer, CE = Cosine Embedding.}
\label{tab:meth-1}
\begin{tabular}{p{3.3em}p{5.7em}p{5.7em}p{5.7em}}
\toprule
\textbf{Loss type} & \textbf{CLIP} & \textbf{Stable \newline Diffusion} & \textbf{VLM (LLaVA)} \\
\midrule
$\mathcal{L}_{\text{retain}}$
& Symmetric InfoNCE contrastive loss~\cite{radford2021clip}
& $\epsilon$-prediction denoising loss~\cite{rombach2022high}
& Symmetric InfoNCE contrastive loss (vision encoder)~\cite{liu2023visual} \\
$\mathcal{L}_{\text{forget}}$
& CE ``push-away'' loss on forget image--text pairs~\cite{cai2025slug}
& CE loss on text encoder for forget prompts~\cite{cai2025slug}
& CE loss on vision encoder for forget concepts~\cite{cai2025slug} \\
$\mathcal{L}_{\text{mismatch}}$
& SDR matching CLIP scores to a frozen baseline \cite{mistretta2025cross}
& SDR on guidance/similarity scores for generations \cite{cong2025guiding}
& Similarity/logit-drift regularizer on neutral QA/prompts \cite{wang2024understanding} \\
\bottomrule
\end{tabular}
\vspace{-0.5cm}
\end{table}

\textbf{CLIP} aims to align visual and textual representations in a shared embedding space. It contains a text encoder and an image encoder, like Vision Transformer (ViT). Vit divides an image into patches and uses a transformer to analyze them.
We compute $g^f_l, g^r_l$ from contrastive losses (Eq.~\ref{eq:retain}, \ref{eq:forget}). Candidate $l$ includes Multi-Head Self-Attention (MSA) heads and Multilayer perceptron (MLP) blocks; vision/text towers are handled jointly.

\textbf{Stable Diffusion} model generates detailed images from text prompts by reversing a noising process.
RAZOR edits the text encoder (CLIP ViT-L/14) that conditions the UNet~\cite{rombach2022high}, following layer-local strategies~\cite{cai2025slug,fan2024salun}. 

\textbf{VLMs} combine computer vision and natural language processing to understand and generate content from both images and text. We modify the vision encoder or its projection to visual tokens to break alignment for forgotten concepts while preserving downstream tasks~\cite{liu2024improved}.

Table~\ref{tab:meth-1} presents the specific retain, forget, and mismatch losses used for each underlying model that we apply within the RAZOR framework. More details can be found in Appendix~\ref{app:theory}.

\textbf{Complexity and Storage.}
RAZOR updates a small subset of layers $\lvert\mathcal{K}\rvert \ll \lvert\mathcal{L}\rvert$.
The required storage is minimal, limited to changed weights (and optionally gradients)  
and significantly more efficient than full fine-tuning or dense masking~\cite{fan2024salun}.

\section{Experimental Setup}
\label{sec:exp_setup}

\noindent
All experiments are conducted on contrastive vision--language encoders, primarily CLIP~\cite{radford2021clip}, using the \texttt{ViT-B/32} and \texttt{ViT-L/14} backbones, inspired by setup in SLUG~\cite{cai2025slug}. The model parameters are sourced from the OpenCLIP repository~\cite{cherti2023openclip}, trained on the LAION-400M dataset~\cite{schuhmann2021laion400m}. Unless otherwise specified, experiments are performed on NVIDIA A100~(40\,GB) GPUs using mixed-precision inference and training.

\begin{table}[t]
\centering
\footnotesize
\renewcommand{\arraystretch}{1.5}
\caption{Unlearning scenarios in RAZOR and the corresponding model families and evaluation metrics.}
\label{tab:unlearning_scenarios}
\begin{tabular}{p{1.5cm}| p{1.7cm} p{1.7cm} p{1.8 cm}}
\toprule
\textbf{Unlearning Scenario} & \textbf{Models} & \textbf{Datasets} & \textbf{Evaluation \newline Metrics} \\ \midrule
\multirow{2}{1.5cm}{\textbf{Identity} \newline \textbf{Removal}} & CLIP & CIFAR-10, LAION-400M, ImageNet-1K, CelebA & M1--M5 \\ \cline{2-4} 
 & VLMs (LLaVA) & CelebA & FA, MME, \newline GQA, \newline MMBench \\ \hline
\textbf{Copyrighted Content \newline Removal} & Stable \newline Diffusion \newline(SD-V1.5,\newline SD-V3) & CelebA & UA, IRA, CRA \newline (UnlearnCanvas) \\ \hline 
\multirow{3}{1.5cm}{\textbf{Concept \& Style Unlearning}} & CLIP & {CelebA} & M1--M5 \\ \cline{2-4} 
 & Stable \newline Diffusion & CelebA & UA, IRA, CRA \\ \cline{2-4} 
 & VLMs & CelebA & FA, MME, \newline GQA, \newline MMBench \\
\bottomrule
\end{tabular}
\vspace{-0.5cm}
\end{table}

\textbf{Unlearning Scenarios.} We consider unlearning scenarios following~\cite{cai2025slug}, and reported in table \ref{tab:unlearning_scenarios}.

\subsection{Datasets and Data Construction}
We use a combination of large-scale public datasets:

\textbf{CIFAR-10}~\cite{krizhevsky2009learning} consists of natural images (32$\times$32 resolution) evenly distributed across ten object categories. It serves as a lightweight yet diverse benchmark for evaluating unlearning performance.

\textbf{LAION-400M}~\cite{schuhmann2021laion400m} 
serves as the primary data source for constructing both $\mathcal{D}_f$ and $\mathcal{D}_r$. The forget set is curated using keyword filters on text captions (e.g., “Elon Musk,” “Taylor Swift”) to collect 4,000 image–text pairs per identity. The retention set uses a random LAION shard ($\sim$7.9K pairs), ensuring that no captions overlap with the forget set.

\textbf{CelebA}~\cite{liu2015celeba} is used for utility evaluation via zero-shot classification on human identities. A sample of 100 frequent identities from LAION overlap is selected for consistency ~\cite{cai2025slug}.

\textbf{ImageNet-1K}~\cite{deng2009imagenet} is used for post-unlearning generalization (utility) measurement. Top-1 accuracy is reported on the validation split to capture any degradation in visual recognition performance.

We adopt \textbf{UnlearnCanvas Benchmark} for style and object unlearning tasks~\cite{zhang2024unlearncanvas}.
UnlearnCanvas is a benchmark designed to evaluate the effectiveness of diffusion models. We select 20 objects and 60 styles, with each forget set containing 400 (style) or 1200 (object) image–text pairs. The retain set reuses LAION shards with neutral content to preserve alignment and global semantics.

\textbf{Implementation Details}
All experiments use the official OpenCLIP tokenizer and cosine-similarity evaluation as in~\cite{radford2021clip}. Following the single-layer or ratio-aware selection principle, we precompute gradients of the forget loss ( $\nabla_\theta\mathcal{L}_{\text{forget}}$) and retain loss( $\nabla_\theta\mathcal{L}_{\text{retain}}$).

\subsection{Evaluation Metrics}
\label{sec:metrics}

Across the different scenarios, we evaluate RAZOR, 
using five quantitative metrics (M1–M5) and three generative alignment measures (UA, IRA, CRA) to assess forgetting precision, privacy safety, and utility preservation, following established unlearning benchmarks~\cite{cai2025slug,fan2024salun,foster2024fast,thudi2022unrolling,mehta2023esd}. All results are reported as mean 
over five random seeds. Arrows (↑ / ↓ / → 0) indicate the desired trend.
Together, these metrics summarize the four key goals of unlearning: forget precision (M1–M2), privacy safety (M3), utility retention (M4–M5), and stylistic coherence (UA–CRA).

\textbf{CLIP} We conduct both identity and object unlearning experiments on CLIP and evaluate performance using the aforementioned quantitative metrics to verify effective forgetting and retention behavior.

\begin{table*}[bt]
\centering
\caption{Comparison of unlearning methods, on CLIP model across CIFAR-10, ImageNet, and LAION-400M datasets on five metrics: M1 (↓): Forget Accuracy, M2 (↓): Forget-Set Cosine Similarity, M3 (→ 0): PrivLeak, M4 (↑): Utility / Preservation, M5 (↑): Retain Retrieval Stability.
}
\resizebox{\textwidth}{!}{
\begin{tabular}{lccccc|ccccc|ccccc}
\toprule
\textbf{Dataset} & \multicolumn{5}{c}{\textbf{CIFAR-10}} & \multicolumn{5}{c}{\textbf{ImageNet}} & \multicolumn{5}{c}{\textbf{LAION-400M}} \\
\cmidrule(lr){2-6} \cmidrule(lr){7-11} \cmidrule(lr){12-16}
\textbf{Method} & M1 ↓ & M2 ↓ & M3 → 0 & M4 ↑ & M5 ↑ & M1 ↓ & M2 ↓ & M3 → 0 & M4 ↑ & M5 ↑ & M1 ↓ & M2 ↓ & M3 → 0 & M4 ↑ & M5 ↑ \\
\midrule
SSD  \cite{foster2024fast} & \textbf{52.00} & 22.42 & 0.40 & 25.00 & 97.50 & \textbf{52.50} & \textbf{24.00} & 0.20 & 30.00 & 97.50 & 42.00 & \textbf{22.00} & 0.10 & 48.00 & 98.00 \\
\rowcolor{lightyellow}
SSD (Quan. 8 bit) & 60.48 & 22.64 & 0.00 & 24.75 & 96.54 & 60.99 & 24.24 & 0.00 & 29.70 & 96.54 & 58.39 & 22.22 & 0.00 & 47.53 & 96.94 \\
\rowcolor{lightpink}
SSD (Quan. 4 bit) & 62.49 & 22.50 & 0.00 & 24.69 & 96.63 & 63.00 & 24.08 & 0.00 & 29.55 & 96.63 & 54.40 & 22.08 & 0.00 & 47.41 & 96.63 \\
LoTUS \cite{spartalis2025lotus}& 54.00 & 24.20 & 1.00 & 35.00 & 75.00 & 54.00 & 23.20 & 1.00 & 36.00 & 79.00 & 44.00 & 23.00 & 0.80 & 56.00 & 82.00 \\
\rowcolor{lightyellow}
LoTUS (Quan. 8 bit) & 62.50 & 26.44 & 0.00 & 34.65 & 74.26 & 63.50 & 23.43 & 0.00 & 35.64 & 78.22 & 54.50 & 23.23 & 0.00 & 55.45 & 81.19 \\
\rowcolor{lightpink}
LoTUS (Quan. 4 bit) & 64.51 & 28.29 & 0.00 & 34.57 & 73.95 & 64.51 & 23.28 & 0.00 & 35.55 & 77.89 & 54.52 & 23.08 & 0.00 & 55.31 & 80.85 \\
SalUn \cite{fan2024salun} & 97.00 & 28.42 & 1.28 & 83.00 & 84.50 & 88.00 & 24.00 & 1.00 & 84.00 & 98.00 & 48.00 & 23.20 & 0.80 & 88.00 & 98.40 \\
\rowcolor{lightyellow}
SalUn (Quan. 8 bit) & 88.60 & 32.70 & 0.00 & 82.18 & 83.67 & 88.81 & 34.24 & 0.00 & 82.17 & 96.03 & 58.89 & 26.43 & 0.00 & 85.13 & 96.37 \\
\rowcolor{lightpink}
SalUn  (Quan. 4 bit) & 88.91 & 34.52 & 0.00 & 81.97 & 83.31 & 98.83 & 24.08 & 0.00 & 82.96 & 96.63 & 58.91 & 26.28 & 0.00 & 86.91 & 96.03 \\
SLUG \cite{cai2025slug}& 67.50 & 27.20 & 1.20 & 87.50 & 96.50 & 68.00 & 28.29 & 0.80 & 88.00 & 99.50 & 48.00 & 28.20 & 0.40 & 88.00 & 99.80 \\
\rowcolor{lightyellow}
SLUG (Quan. 8 bit) & 68.12 & 27.47 & 0.00 & 86.63 & 95.54 & 68.63 & 28.57 & 0.00 & 87.13 & 98.57 & 48.44 & 28.48 & 0.00 & 87.13 & 98.81 \\
\rowcolor{lightpink}
SLUG (Quan. 4 bit) & 68.13 & 27.35 & 0.00 & 86.41 & 95.12 & 68.64 & 28.39 & 0.00 & 86.91 & 98.12 & 48.46 & 28.30 & 0.00 & 86.91 & 98.37 \\
\textbf{RAZOR (Ours)} & 52.50 & \textbf{22.00} & \textbf{0.00} & \textbf{89.00} & \textbf{100.00} & 53.50 & 27.51 & \textbf{0.00} & \textbf{92.00} & \textbf{100.00} & \textbf{40.00} & 27.46 & \textbf{0.00} & \textbf{94.00} & \textbf{100.00} \\
\rowcolor{lightyellow}
RAZOR (Quan. 8 bit) & 52.99 & 22.22 & 0.00 & 88.12 & 99.01 & 54.00 & 27.78 & 0.00 & 91.09 & 99.01 & 40.37 & 27.73 & 0.00 & 93.07 & 99.01 \\
\rowcolor{lightpink}
RAZOR (Quan. 4 bit) & 53.00 & 22.08 & 0.00 & 87.89 & 98.60 & 54.01 & 27.61 & 0.00 & 90.86 & 98.60 & 40.38 & 27.60 & 0.00 & 92.83 & 98.60 \\
\bottomrule
\end{tabular}
}
\label{tab:full_results_quantized}
\end{table*}

\textbf{(M1 ↓) Forget Accuracy / Verbatim Memorization.}
Measures residual recognition of forgotten concepts by computing CLIP logits for forget-set images against class prompts $[q_{\text{forget}}, q_{\text{retain}}]$.  
Lower scores denote effective suppression of the forgotten class~\cite{cai2025slug}.
\textbf{(M2 ↓) Forget-Set Cosine Similarity.}
Evaluates semantic alignment between forget-set images and their text prompts.  
Reduced similarity reflects diminished memorized knowledge~\cite{fan2024salun,foster2024fast}.
\textbf{(M3 → 0) Privacy Leakage.}
Quantifies residual privacy exposure as mean-squared drift of image–caption similarities before and after unlearning.  
Values approaching zero indicate minimal private information retention~\cite{thudi2022unrolling,mehta2023esd}.
\textbf{(M4 ↑) Retain Accuracy / Utility Preservation.}
Zero-shot accuracy on retain-set images using dual prompts.  
Higher values imply strong utility and stable representations~\cite{cai2025slug,fan2024salun}.
\textbf{(M5 ↑) Retain Retrieval Stability.}
Defined as 
$\text{M5} = 1 - |\text{Util}_{\text{after}} - \text{Util}_{\text{before}}|$,  
it measures consistency of retrieval accuracy; values near 1.0 signify minimal collateral drift~\cite{siddiqui2025from}.

\noindent \textbf{Model Quantization.}
To assess RAZOR’s robustness under low-precision conditions, we apply post-training quantization to the CLIP model, following concerns raised by Zhang et al.~\cite{zhang2024catastrophic} regarding quantization-induced unlearning failure. Model weights are converted to 8-bit and 4-bit formats, significantly reducing memory and computation costs while enabling analysis of the trade-off between unlearning accuracy and efficiency~\cite{gholami2022survey,frantar2022gptq}.

\noindent \textbf{Stable Diffusion models.}
We conduct both object and style unlearning on SD-v1.5 and SD-v3.1 models, evaluating performance using the aforementioned metrics to measure forgetting effectiveness and generative fidelity.

\noindent \textbf{(UA ↑) Unlearned Accuracy.}
Assesses generative models’ ability to reproduce forgotten styles (e.g., “pixel art”) only when explicitly prompted~\cite{mehta2023esd}.
\textbf{(IRA ↑) In-Retain Alignment.}
Measures stylistic coherence of generated retain-style images, reflecting robustness and preserved alignment~\cite{rombach2022high}.
\textbf{(CRA ↑) Content/Robustness Alignment.}
Evaluates neutral-style generations for semantic consistency and robustness against over-forgetting~\cite{rombach2022high,liu2024improved}.

\begin{table*}[tb]
\centering
\caption{Comparison of unlearning performance on Stable Diffusion SD-V3 and SD-V1.5 across style and object forgetting tasks. Compares different unlearning methods on UnlearnCanvas benchmark, with LION-400M as primary dataset. Metrics: UA (↑): Unlearning Achievement, IRA (↑): In-Retain Alignment, CRA (↑): Content/Robustness Alignment.}
\scriptsize
\resizebox{\textwidth}{!}{
\begin{tabular}{l|ccc|ccc|ccc|ccc}
\toprule
\multirow{2}{*}{\textbf{Method}} 
& \multicolumn{6}{c|}{\textbf{SD-V3}} 
& \multicolumn{6}{c}{\textbf{SD-V1.5}} \\
\cmidrule(lr){2-7} \cmidrule(lr){8-13}
& \multicolumn{3}{c|}{Style} & \multicolumn{3}{c|}{Object} 
& \multicolumn{3}{c|}{Style} & \multicolumn{3}{c}{Object} \\
& UA ↑ & IRA ↑ & CRA ↑ & UA ↑ & IRA ↑ & CRA ↑ 
& UA ↑ & IRA ↑ & CRA ↑ & UA ↑ & IRA ↑ & CRA ↑ \\
\midrule
ESD \cite{gandikota2023erasing} & \textbf{99.62} & 89.97 & 98.86 & 97.44 & 68.47 & 82.37 & 98.58 & 80.97 & 93.96 & 92.15 & 55.78 & 44.23 \\
FMN \cite{zhang2024forget} & 94.58 & 72.97 & 62.96 & 85.15 & 94.50 & 86.23 & 88.48 & 56.77 & 46.60 & 45.64 & 90.63 & 73.46 \\
UCE \cite{gandikota2024unified}  & 98.98 & 72.27 & 64.96 & 96.64 & 56.63 & 64.46 & 98.40 & 60.22 & 47.71 & 94.31 & 39.35 & 34.67 \\
EDiff \cite{xie2025ec}   & 94.40 & 84.22 & 99.23 & 90.06 & 95.07 & 68.67 & 92.42 & 73.91 & 98.93 & 86.67 & 94.03 & 48.48 \\
SalUn  \cite{fan2023salun}  & 90.36 & 92.33 & 97.02 & 91.06 & 98.35 & 99.59 & 86.26 & 90.39 & 95.08 & 86.91 & 96.35 & 99.59 \\
SLUG  \cite{cai2025slug}    & 88.20 & 85.59 & 91.00 & 85.44 & 79.50 & 91.00 & 86.29 & 84.59 & 88.43 & 75.43 & 77.50 & 81.18 \\
\textbf{RAZOR} & 99.40 & \textbf{98.97} & \textbf{100.00} & \textbf{98.80} & \textbf{98.35} & \textbf{100.00} & \textbf{99.26} & \textbf{96.39} & \textbf{99.00} & \textbf{97.91} & \textbf{97.50} & \textbf{99.00} \\
\bottomrule
\end{tabular}
}
\vspace{-0.3cm}
\label{tab:sd_accuracy}
\end{table*}

\textbf{Vision-Language Models (VLMs).}
For vision-language evaluation, we employ LLaVA-v1.6~\cite{liu2024improved} as the primary model. We perform concept erasure and structured unlearning to assess both unlearning effectiveness and utility preservation, evaluated through deletion-oriented and comprehension-based benchmarks.

\textbf{Forget Accuracy (FA)}, similar to M1, measures the residual ability of the model to correctly recognize or reproduce the forgotten concept, where lower values indicate more effective erasure~\cite{cai2025slug}. 
To assess general utility retention, we employ three multimodal benchmarks: 
\textbf{MME}~\cite{fu2023mme}, which captures both perceptual understanding (object, color, and scene recognition) and cognitive reasoning (spatial and commonsense inference); 
\textbf{GQA}~\cite{hudson2019gqa}, which evaluates fine-grained visual reasoning through structured question answering; 
and \textbf{MMBench}~\cite{liu2025mmbench}, a comprehensive suite testing instruction-following, OCR, and semantic grounding capabilities in vision–language models. For additional details, please refer to Appendix~\ref{app:experimental}.

\section{Results}
\label{sec:result}
\vspace{-0.3cm}
We use three different scenarios to evaluate RAZOR, encompassing a variety of tasks, base models, and datasets. For a comprehensive discussion, see Appendix-\ref{app:result}.

\subsection{Results with CLIP} We evaluate how effectively the CLIP model supports targeted identity and style unlearning tasks. Table~\ref{tab:full_results_quantized} compares RAZOR against recent unlearning approaches and shows that it delivers more precise forgetting (M1--M2), perfect M3 accuracy, while simultaneously preserving higher retention and utility (M4--M5). While SSD~\cite{foster2024fast} attains competitive M1 and M2 values in a few settings, RAZOR achieves superior overall accuracy and stability across all five metrics. Importantly, these advantages persist across datasets and model families, including CIFAR-10, ImageNet, and LAION-400M, highlighting RAZOR’s robustness and generalizability as a targeted unlearning framework.

\noindent An important observation from Table~\ref{tab:full_results_quantized} is the clear degradation introduced by post-training quantization: the 8-bit models (yellow rows) exhibit reduced performance relative to full-precision, and the 4-bit models (light red rows) show the most severe decline. Notably, methods that update the \emph{entire} model suffer disproportionately under quantization, whereas approaches that restrict updates to carefully selected layers display substantially greater robustness. These findings align with the observations of Zhang et al.~\cite{zhang2024catastrophic}, who reported significant vulnerability of unlearning methods to quantization-induced degradation.

\noindent \textbf{Qualitatively,} we observe that RAZOR produces clean and stable separation between forgotten and retained concepts, with minimal drift in global embedding structure. Unlike full-model update methods, which often introduce noisy cross-concept interference, RAZOR preserves the semantic geometry of non-target classes, leading to more visually consistent and interpretable unlearning behavior across CLIP evaluations.

\subsection{Results with Stable Diffusion} 
We assess the diffusion model’s ability to perform targeted content and style unlearning. Table~\ref{tab:sd_accuracy} reports a comparison of RAZOR with recent unlearning approaches on the UnlearnCanvas benchmark for both SD-V3 and SD-V1.5. With the improved architecture of SD-V3, all methods exhibit higher absolute performance; however, RAZOR maintains a clear advantage across the majority of metrics. Although ESD~\cite{gandikota2023erasing} marginally outperforms RAZOR on a single metric, RAZOR consistently achieves the strongest overall forgetting fidelity and alignment scores (UA, IRA, CRA) across both diffusion models. These results underscore RAZOR’s robustness, architectural generality, and effectiveness in diffusion-based unlearning settings.

\begin{figure}[bt]
   \includegraphics[width=\linewidth]{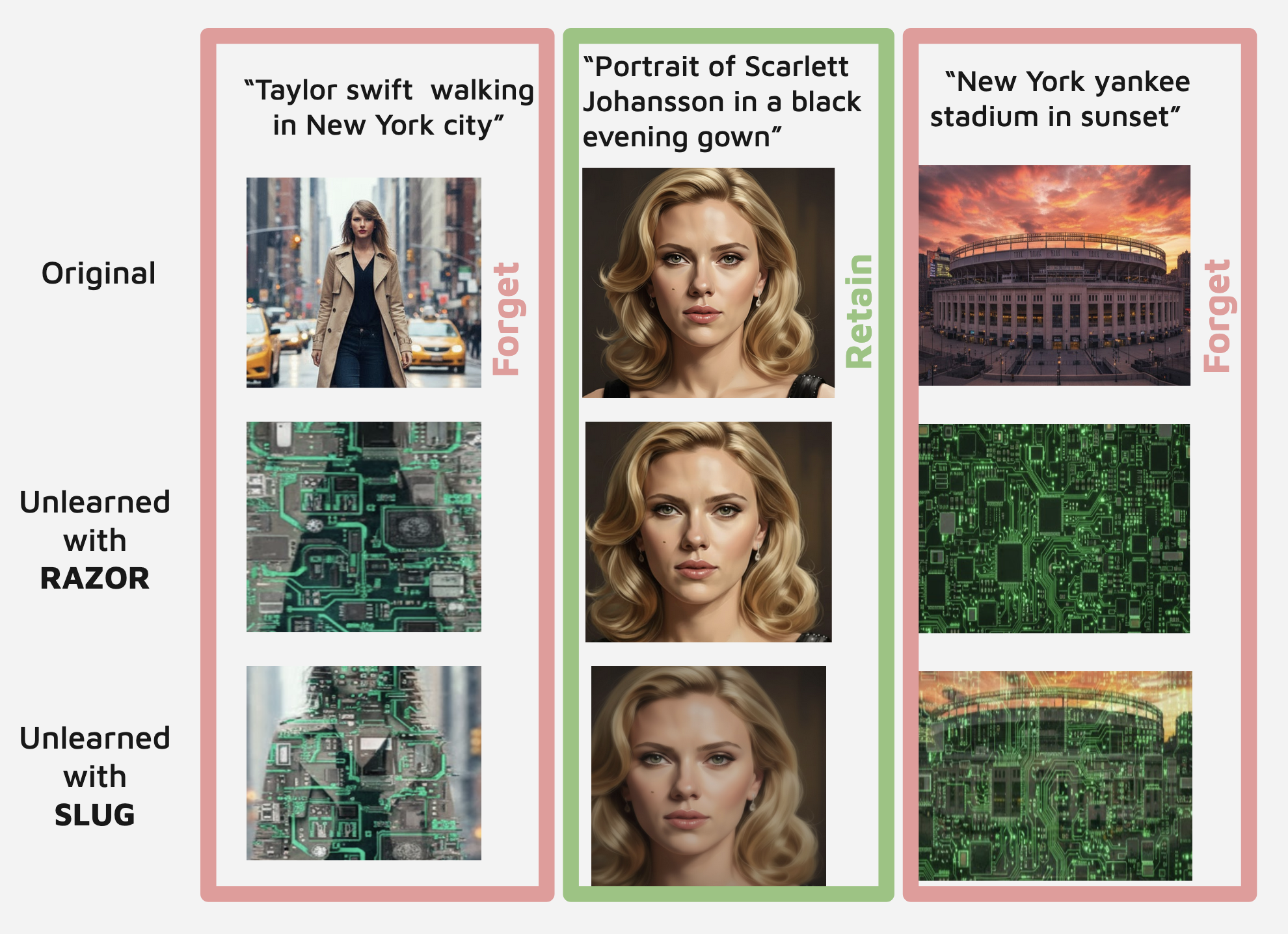}
   \caption{Qualitative comparison of RAZOR and SLUG on SD-3 identity unlearning, showing stronger forgetting with better retention preservation.}
   \label{fig:sd}
\end{figure}

\begin{table}[bt]
\centering
\caption{Efficiency comparison on SD-V1.5 in terms of Time (s), Memory (GB), Storage (GB), and Trade-off. Trade-off is calculated as average accuracy (UA+IRA+CRA over style and object) divided by average cost (time, memory, storage).}
\small
\begin{tabular}{lrrrr}
\toprule
\textbf{Method} & \textbf{Time} ↓ & \textbf{Mem ↓} & \textbf{Storage ↓} & \textbf{Trade-off ↑} \\
\midrule
ESD \cite{gandikota2023erasing}  & 6163  & 17.8  & 4.30  & 11.97 \\
FMN \cite{zhang2024forget} & 350   & 17.9  & 4.20  & 17.31 \\
UCE \cite{gandikota2024unified}     & 434   & 5.1   & 1.70  & 26.78 \\
EDiff \cite{xie2025ec} & 1567  & 27.8  & 4.00  & 11.16 \\
SalUn \cite{fan2024salun}& 667   & 30.8  & 4.00  & 13.88 \\
SLUG \cite{cai2025slug} & \textbf{39}    & \textbf{3.6}  & \textbf{0.04} & 59.42 \\
\textbf{RAZOR} & 78 & 4.2 & 0.06 & \textbf{66.86} \\
\bottomrule
\end{tabular}
\label{tab:sd_efficiency}
\vspace{-0.6cm}
\end{table}
As shown in the figure \ref{fig:sd}, RAZOR cleanly removes the forgotten concept while maintaining high quality retention outputs, whereas SLUG introduces noticeable drift into both forgotten and retained generations.

\noindent \textbf{Efficiency.} Table~\ref{tab:sd_efficiency} summarizes efficiency results on SD-V1.5, including runtime, memory footprint, storage requirements, and a combined accuracy efficiency trade-off metric. While SLUG~\cite{cai2025slug} attains strong performance on certain individual efficiency measures, RAZOR achieves the highest overall trade-off score by jointly delivering strong unlearning fidelity with low computational and memory overhead. This balance allows RAZOR to outperform all baselines by a substantial margin, highlighting its practicality for scalable diffusion-model unlearning. 

\subsection{Results with VLMs}
\begin{figure}[ht!]
   \includegraphics[width=\linewidth]{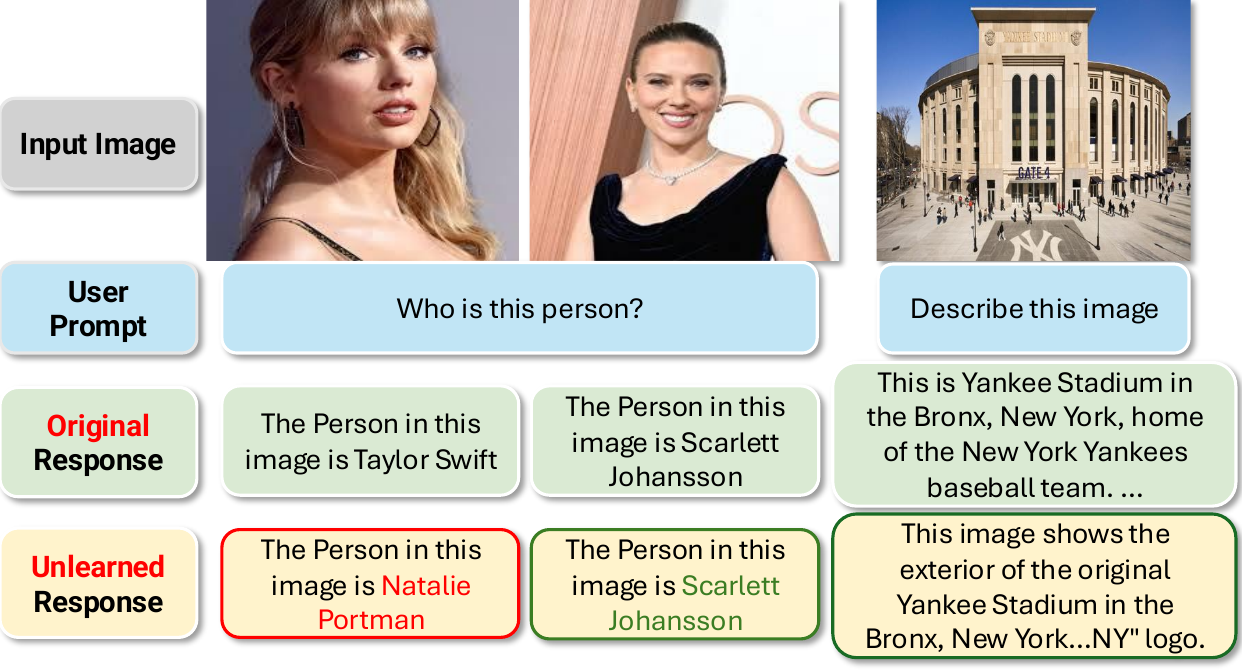}
   \caption{Qualitative example of Identity unlearning on LLaVA-v1.6 with RAZOR shows that ``Taylor Swift'' was successfully unlearned without affecting any other identities/concepts.}
   \label{fig:vlm}
\end{figure}

We evaluate identity and style-level unlearning capabilities in the vision-language model. As shown in Table~\ref{tab:razor_llava}, RAZOR achieves substantial forgetting of targeted identities in LLaVA-1.6-8B, reducing average forget accuracy to only 2.2\% while maintaining near-original performance across cognition, perception, and QA benchmarks. This demonstrates RAZOR’s ability to perform precise, ratio-aware layer edits that effectively erase targeted concepts with minimal collateral degradation of overall utility.
Figure \ref{fig:vlm} demonstrates that RAZOR successfully erases the forgotten identity in vision–language responses while keeping correct recognition and description for all retained concepts.

\begin{table}[t]
\centering
\scriptsize
\caption{Quantitative evaluation of RAZOR on LLaVA-1.6-8B \cite{liu2024improved} for identity unlearning, where higher value is better for each metrics except FA.}
\label{tab:razor_llava}
\setlength{\tabcolsep}{2pt}
\renewcommand{\arraystretch}{0.9}
\begin{tabular}{l|c|ccc}
\toprule
\textbf{Identity to Forget} & 
\textbf{~~FA ↓~~} & 
\textbf{MME (Cogn./Perc.)} & 
\textbf{GQA} & 
\textbf{MMBench (\%)} \\
\midrule
Base Model & 97.25 & 321.5 / 1454.4 & 60.18 & 61.57 \\
\midrule
Scarlett Johansson & 2.0 & 301.4 / 1362.7 & 60.83 & 61.90 \\
Taylor Swift & 1.5 & 339.3 / 1347.9 & 60.86 & 60.77 \\
Robert Downey Jr. & 5.0 & 348.3 / 1217.2 & 58.85 & 56.89 \\
Jeff Bezos & 2.0 & 319.9 / 1328.7 & 60.55 & 61.72 \\
Kanye West & 3.0 & 319.1 / 1378.4 & 61.36 & 61.90 \\
Tom Cruise & 0.0 & 354.8 / 1428.6 & 61.25 & 62.14 \\
Kim Kardashian & 4.0 & 292.7 / 1281.8 & 60.57 & 60.72 \\
Barack Obama & 0.0 & 293.6 / 1302.5 & 60.82 & 61.44 \\
Lady Gaga & 2.0 & 274.9 / 1222.4 & 58.85 & 56.20 \\
Natalie Portman & 0.0 & 328.7 / 1315.2 & 60.63 & 60.52 \\
\midrule
\textbf{Average} & $\mathbf{2.2}$ & 
$\mathbf{317.6 / 1318.9}$ & 
$\mathbf{60.46}$ & 
$\mathbf{60.9}$ \\
\bottomrule
\end{tabular}
\end{table}

\subsection{Ablation Study}
\label{sec:ablation}
We conduct targeted ablations to isolate the contribution of RAZOR's design components, including the multi-objective loss and layer selection strategy. As shown in Table~\ref{tab:ablation},
removing any one component or applying losses without selection or refinement leads to degraded alignment or unintended forgetting, confirming the importance of each component. As expected, removing the forget loss substantially degrades unlearning performance, confirming its central role in driving concept erasure. We also observe that updating all layers (row 4) results in higher forgetting accuracy (i.e., lower M1, M2) compared to our proposed selective strategy, since full-model updates aggressively suppress the target concept across the entire network. However, this brute-force approach comes at a significant cost: it produces the worst outcomes across all remaining metrics, including privacy leakage (M3), utility retention (M4), and retrieval stability (M5). These results clearly indicate that RAZOR’s combination of ratio-aware selection, multi-objective loss, and iterative refinement provides the most balanced and effective unlearning behavior. Further ablation details are provided in Appendix~\ref{app:ablation}.

\begin{table}[t]
\centering
\caption{Ablation study on loss components and update strategies using CLIP on CIFAR-10 dataset. M1: Forget Accuracy, M2: Cosine Similarity, M3: PrivLeak, M4: Utility, M5: Retain Stability.}
\resizebox{0.48\textwidth}{!}{
\begin{tabular}{lccccr}
\toprule
\textbf{Selection} & \textbf{M1} ↓ & \textbf{M2} ↓ & \textbf{M3}→0 & \textbf{M4} ↑ & \textbf{M5} ↑ \\
\midrule
Losses w/o Retain     & 52.72 & 23.00 & 0.40 & 82.00 & 99.0 \\
Losses w/o Mismatch     & 53.25 & 27.04 & 1.04 & 88.00 & 100.0 \\
Losses w/o Forget     & 96.00 & 28.42 & 0.40 & 86.00 & 99.0 \\
All Losses, No Selection     & 51.00 & 21.04 & 1.58 & 78.00 & 96.0 \\
All Losses, No Iterations     & 53.00 & 22.00 & 0.00 & 88.82 & 100.0 \\
Full (All Losses + Iteration) & \textbf{52.50} & \textbf{22.00} & \textbf{0.00} & \textbf{89.00} & \textbf{100.0} \\
\bottomrule
\end{tabular}
}
\label{tab:ablation}
\vspace{-1em}
\end{table}

\section{Conclusion}
\label{sec:conclusion}
RAZOR introduces a ratio-aware, multi-layer unlearning framework that achieves precise knowledge removal without compromising model utility or scalability. By combining targeted layer-head selection, a multi-objective loss, and iterative refinement, RAZOR enables reliable unlearning across both discriminative and generative models, including CLIP, stable diffusion, and vision–language systems. Our results demonstrate that 
careful minimal edits can match or outperform full-model updates, even after quantization, highlighting a path toward practical, real-time unlearning in deployed systems. This work challenges the assumption that precision and efficiency must trade off, and establishes a blueprint for scalable model editing. 
Future work should explore robustness under adversarial unlearning, cross-task generalization, and broader application domains such as audio and video generation.

{
    \small
    \bibliographystyle{ieeenat_fullname}
    \bibliography{main}

@String(CVPR= {IEEE Conf. Comput. Vis. Pattern Recog.})

@String(ICCV= {Int. Conf. Comput. Vis.})

@String(AAAI = {AAAI})

@String(CVPR  = {CVPR})

@String(ICCV  = {ICCV})

@inproceedings{cai2025slug,
  title     = {Targeted Unlearning with Single Layer Unlearning Gradient},
  author    = {Cai, Zikui and Tan, Yaoteng and Asif, M. Salman},
  booktitle = {Proceedings of the 42nd International Conference on Machine Learning},
  year      = {2025},
  publisher = {PMLR}
}

@inproceedings{fan2024salun,
  title     = {SalUn: Empowering Machine Unlearning via Gradient-based Weight Saliency in Both Image Classification and Generation},
  author    = {Fan, Chengzhi and Liu, Jiawei and Zhang, Yiding and Wei, Ding and Wong, Eric and Liu, Sijia},
  booktitle = {International Conference on Learning Representations},
  year      = {2024},
  eprint    = {2310.12508},
  archivePrefix = {arXiv}
}

@inproceedings{foster2024fast,
  title={Fast machine unlearning without retraining through selective synaptic dampening},
  author={Foster, Jack and Schoepf, Stefan and Brintrup, Alexandra},
  booktitle={Proceedings of the AAAI conference on artificial intelligence},
  volume={38},
  number={11},
  pages={12043--12051},
  year={2024}
}

@inproceedings{radford2021clip,
  title     = {Learning Transferable Visual Models From Natural Language Supervision},
  author    = {Radford, Alec and Kim, Jong Wook and Hallacy, Chris and Ramesh, Aditya and Goh, Gabriel and Agarwal, Sandhini and Sastry, Girish and Askell, Amanda and Mishkin, Pamila and Clark, Jack and others},
  booktitle = {International Conference on Machine Learning},
  year      = {2021},
  pages     = {8748--8763},
  publisher = {PMLR},
  eprint    = {2103.00020},
  archivePrefix = {arXiv}
}

@article{schuhmann2021laion400m,
  title   = {LAION-400M: Open Dataset of CLIP-filtered 400 Million Image-Text Pairs},
  author  = {Schuhmann, Christoph and Vencu, Richard and Beaumont, Romain and Kaczmarczyk, Robert and Mullis, Clayton and Katta, Aarush and Coombes, Theo and Jitsev, J{\"o}rn and Komatsuzaki, Aran},
  journal = {arXiv preprint arXiv:2111.02114},
  year    = {2021}
}

@article{cherti2023openclip,
  title   = {Reproducible Scaling Laws for Contrastive Language–Image Learning},
  author  = {Cherti, Mehdi and Beaumont, Romain and Wightman, Ross and Wortsman, Mitchell and Ilharco, Gabriel and Schmidt, Ludwig and Jitsev, J{\"o}rn},
  journal = {arXiv preprint arXiv:2309.16671},
  year    = {2023}
}

@article{shah2023unlearning,
  title={Unlearning via sparse representations},
  author={Shah, Vedant and Tr{\"a}uble, Frederik and Malik, Ashish and Larochelle, Hugo and Mozer, Michael and Arora, Sanjeev and Bengio, Yoshua and Goyal, Anirudh},
  journal={arXiv preprint arXiv:2311.15268},
  year={2023}
}

@inproceedings{golatkar2020forgetting,
  title={Forgetting outside the box: Scrubbing deep networks of information accessible from input-output observations},
  author={Golatkar, Aditya and Achille, Alessandro and Soatto, Stefano},
  booktitle={European Conference on Computer Vision},
  pages={383--398},
  year={2020},
  organization={Springer}
}

@article{kurmanji2023towards,
  title={Towards unbounded machine unlearning},
  author={Kurmanji, Meghdad and Triantafillou, Peter and Hayes, Jamie and Triantafillou, Eleni},
  journal={Advances in neural information processing systems},
  volume={36},
  pages={1957--1987},
  year={2023}
}

@inproceedings{zhang2024forget,
  title={Forget-me-not: Learning to forget in text-to-image diffusion models},
  author={Zhang, Gong and Wang, Kai and Xu, Xingqian and Wang, Zhangyang and Shi, Humphrey},
  booktitle={Proceedings of the IEEE/CVF conference on computer vision and pattern recognition},
  pages={1755--1764},
  year={2024}
}

@inproceedings{wu2025erasing,
  title={Erasing undesirable influence in diffusion models},
  author={Wu, Jing and Le, Trung and Hayat, Munawar and Harandi, Mehrtash},
  booktitle={Proceedings of the Computer Vision and Pattern Recognition Conference},
  pages={28263--28273},
  year={2025}
}

@inproceedings{liu2015celeba,
  title     = {Deep Learning Face Attributes in the Wild},
  author    = {Liu, Ziwei and Luo, Ping and Wang, Xiaogang and Tang, Xiaoou},
  booktitle = {Proceedings of the IEEE International Conference on Computer Vision (ICCV)},
  pages     = {3730--3738},
  year      = {2015}
}

@inproceedings{deng2009imagenet,
  title     = {ImageNet: A Large-Scale Hierarchical Image Database},
  author    = {Deng, Jia and Dong, Wei and Socher, Richard and Li, Li-Jia and Li, Kai and Fei-Fei, Li},
  booktitle = {Proceedings of the IEEE Conference on Computer Vision and Pattern Recognition (CVPR)},
  pages     = {248--255},
  year      = {2009}
}

@inproceedings{cao2015towards,
  title={Towards making systems forget with machine unlearning},
  author={Cao, Yinzhi and Yang, Junfeng},
  booktitle={2015 IEEE symposium on security and privacy},
  pages={463--480},
  year={2015},
  organization={IEEE}
}

@inproceedings{bourtoule2021machine,
  title={Machine unlearning},
  author={Bourtoule, Lucas and Chandrasekaran, Varun and Choquette-Choo, Christopher A and Jia, Hengrui and Travers, Adelin and Zhang, Baiwu and Lie, David and Papernot, Nicolas},
  booktitle={2021 IEEE symposium on security and privacy (SP)},
  pages={141--159},
  year={2021},
  organization={IEEE}
}

@inproceedings{wu2024scissorhands,
  title={Scissorhands: Scrub data influence via connection sensitivity in networks},
  author={Wu, Jing and Harandi, Mehrtash},
  booktitle={European Conference on Computer Vision},
  pages={367--384},
  year={2024},
  organization={Springer}
}

@article{foster2024loss,
  title={Loss-free machine unlearning},
  author={Foster, Jack and Schoepf, Stefan and Brintrup, Alexandra},
  journal={arXiv preprint arXiv:2402.19308},
  year={2024}
}

@article{zarsky2016incompatible,
  title={Incompatible: The GDPR in the age of big data},
  author={Zarsky, Tal Z},
  journal={Seton Hall L. Rev.},
  volume={47},
  pages={995},
  year={2016},
  publisher={HeinOnline}
}

@inproceedings{patel2025learning,
  title={Learning to unlearn while retaining: Combating gradient conflicts in machine unlearning},
  author={Patel, Gaurav and Qiu, Qiang},
  booktitle={Proceedings of the IEEE/CVF International Conference on Computer Vision},
  pages={4211--4221},
  year={2025}
}

@article{kyu2024contrastive,
  title={Contrastive unlearning: A contrastive approach to machine unlearning},
  author={kyu Lee, Hong and Zhang, Qiuchen and Yang, Carl and Lou, Jian and Xiong, Li},
  journal={arXiv},
  year={2024}
}

@inproceedings{spartalis2025lotus,
  title={LoTUS: Large-Scale Machine Unlearning with a Taste of Uncertainty},
  author={Spartalis, Christoforos N and Semertzidis, Theodoros and Gavves, Efstratios and Daras, Petros},
  booktitle={Proceedings of the Computer Vision and Pattern Recognition Conference},
  pages={10046--10055},
  year={2025}
}

@inproceedings{gandikota2023erasing,
  title={Erasing concepts from diffusion models},
  author={Gandikota, Rohit and Materzynska, Joanna and Fiotto-Kaufman, Jaden and Bau, David},
  booktitle={Proceedings of the IEEE/CVF international conference on computer vision},
  pages={2426--2436},
  year={2023}
}

@article{cywinski2025saeuron,
  title={SAeUron: Interpretable concept unlearning in diffusion models with sparse autoencoders},
  author={Cywi{\'n}ski, Bartosz and Deja, Kamil},
  journal={arXiv preprint arXiv:2501.18052},
  year={2025}
}

@inproceedings{xiao2025reminiscence,
  title={Reminiscence attack on residuals: Exploiting approximate machine unlearning for privacy},
  author={Xiao, Yaxin and Ye, Qingqing and Hu, Li and Zheng, Huadi and Hu, Haibo and Liang, Zi and Li, Haoyang and Jiao, Yijie},
  booktitle={Proceedings of the IEEE/CVF International Conference on Computer Vision},
  pages={3058--3068},
  year={2025}
}

@inproceedings{rombach2022high,
  title={High-resolution image synthesis with latent diffusion models},
  author={Rombach, Robin and Blattmann, Andreas and Lorenz, Dominik and Esser, Patrick and Ommer, Bj{\"o}rn},
  booktitle={Proceedings of the IEEE/CVF Conference on Computer Vision and Pattern Recognition},
  pages={10684--10695},
  year={2022}
}

@inproceedings{liu2024improved,
  title={Improved baselines with visual instruction tuning},
  author={Liu, Haotian and Li, Chunyuan and Li, Yuheng and Lee, Yong Jae},
  booktitle={Proceedings of the IEEE/CVF Conference on Computer Vision and Pattern Recognition},
  year={2024}
}

@article{nguyen2022survey,
  title={A survey of machine unlearning},
  author={Nguyen, Tuan Tuan and Zhou, Renjie and Luo, Pan and Jiang, Jing and others},
  journal={arXiv preprint arXiv:2209.02299},
  year={2022}
}

@article{carmi2023european,
  title={The European General Data Protection Regulation (GDPR) in mHealth: Theoretical and practical aspects for practitioners’ use},
  author={Carmi, Lior and Zohar, Mishael and Riva, Gianluigi M},
  journal={Medicine, Science and the Law},
  volume={63},
  number={1},
  pages={61--68},
  year={2023},
  publisher={SAGE Publications Sage UK: London, England}
}

@article{jia2023model,
  title={Model sparsity can simplify machine unlearning},
  author={Jia, Jinghan and Liu, Jiancheng and Ram, Parikshit and Yao, Yuguang and Liu, Gaowen and Liu, Yang and Sharma, Pranay and Liu, Sijia},
  journal={Advances in Neural Information Processing Systems},
  volume={36},
  pages={51584--51605},
  year={2023}
}

@misc{triantafillou2023evaluation,
  title={Evaluation for the NeurIPS Machine Unlearning Competition},
  author={Triantafillou, Eleni and Kairouz, Peter},
  year={2023},
  publisher={aug}
}

@article{goel2022towards,
  title={Towards adversarial evaluations for inexact machine unlearning},
  author={Goel, Shashwat and Prabhu, Ameya and Sanyal, Amartya and Lim, Ser-Nam and Torr, Philip and Kumaraguru, Ponnurangam},
  journal={arXiv preprint arXiv:2201.06640},
  year={2022}
}

@inproceedings{wu2023certified,
  title={Certified edge unlearning for graph neural networks},
  author={Wu, Kun and Shen, Jie and Ning, Yue and Wang, Ting and Wang, Wendy Hui},
  booktitle={Proceedings of the 29th ACM SIGKDD Conference on Knowledge Discovery and Data Mining},
  pages={2606--2617},
  year={2023}
}

@article{shaik2024exploring,
  title={Exploring the landscape of machine unlearning: A comprehensive survey and taxonomy},
  author={Shaik, Thanveer and Tao, Xiaohui and Xie, Haoran and Li, Lin and Zhu, Xiaofeng and Li, Qing},
  journal={IEEE Transactions on Neural Networks and Learning Systems},
  year={2024},
  publisher={IEEE}
}

@inproceedings{lyu2024one,
  title={One-dimensional adapter to rule them all: Concepts diffusion models and erasing applications},
  author={Lyu, Mengyao and Yang, Yuhong and Hong, Haiwen and Chen, Hui and Jin, Xuan and He, Yuan and Xue, Hui and Han, Jungong and Ding, Guiguang},
  booktitle={Proceedings of the IEEE/CVF Conference on Computer Vision and Pattern Recognition},
  pages={7559--7568},
  year={2024}
}

@article{yao2024large,
  title={Large language model unlearning},
  author={Yao, Yuanshun and Xu, Xiaojun and Liu, Yang},
  journal={Advances in Neural Information Processing Systems},
  volume={37},
  pages={105425--105475},
  year={2024}
}

@misc{guo2023certifieddataremovalmachine,
      title={Certified Data Removal from Machine Learning Models}, 
      author={Chuan Guo and Tom Goldstein and Awni Hannun and Laurens van der Maaten},
      year={2023},
      eprint={1911.03030},
      archivePrefix={arXiv},
      primaryClass={cs.LG},
      url={https://arxiv.org/abs/1911.03030}, 
}

@article{cottier2024rising,
  title={The rising costs of training frontier AI models},
  author={Cottier, Ben and Rahman, Robi and Fattorini, Loredana and Maslej, Nestor and Besiroglu, Tamay and Owen, David},
  journal={arXiv preprint arXiv:2405.21015},
  year={2024}
}

@article{guerra2023cost,
  title={The cost of training machine learning models over distributed data sources},
  author={Guerra, Elia and Wilhelmi, Francesc and Miozzo, Marco and Dini, Paolo},
  journal={IEEE Open Journal of the Communications Society},
  volume={4},
  pages={1111--1126},
  year={2023},
  publisher={IEEE}
}

@article{ginart2019making,
  title={Making ai forget you: Data deletion in machine learning},
  author={Ginart, Antonio and Guan, Melody and Valiant, Gregory and Zou, James Y},
  journal={Advances in neural information processing systems},
  volume={32},
  year={2019}
}

@article{nguyen2025survey,
  title={A survey of machine unlearning},
  author={Nguyen, Thanh Tam and Huynh, Thanh Trung and Ren, Zhao and Nguyen, Phi Le and Liew, Alan Wee-Chung and Yin, Hongzhi and Nguyen, Quoc Viet Hung},
  journal={ACM Transactions on Intelligent Systems and Technology},
  volume={16},
  number={5},
  pages={1--46},
  year={2025},
  publisher={ACM New York, NY}
}

@article{zhang2023review,
  title={A review on machine unlearning},
  author={Zhang, Haibo and Nakamura, Toru and Isohara, Takamasa and Sakurai, Kouichi},
  journal={SN Computer Science},
  volume={4},
  number={4},
  pages={337},
  year={2023},
  publisher={Springer}
}

@article{wang2023machine,
  title={Machine unlearning via representation forgetting with parameter self-sharing},
  author={Wang, Weiqi and Zhang, Chenhan and Tian, Zhiyi and Yu, Shui},
  journal={IEEE Transactions on Information Forensics and Security},
  volume={19},
  pages={1099--1111},
  year={2023},
  publisher={IEEE}
}

@article{martens2020new,
  title={New insights and perspectives on the natural gradient method},
  author={Martens, James},
  journal={Journal of Machine Learning Research},
  volume={21},
  number={146},
  pages={1--76},
  year={2020}
}

@article{yu2020gradient,
  title={Gradient surgery for multi-task learning},
  author={Yu, Tianhe and Kumar, Saurabh and Gupta, Abhishek and Levine, Sergey and Hausman, Karol and Finn, Chelsea},
  journal={Advances in neural information processing systems},
  volume={33},
  pages={5824--5836},
  year={2020}
}

@inproceedings{thudi2022unrolling,
  title={Unrolling sgd: Understanding factors influencing machine unlearning},
  author={Thudi, Anvith and Deza, Gabriel and Chandrasekaran, Varun and Papernot, Nicolas},
  booktitle={2022 IEEE 7th European Symposium on Security and Privacy (EuroS\&P)},
  pages={303--319},
  year={2022},
  organization={IEEE}
}

@InProceedings{Hoang_2024_WACV,
    author    = {Hoang, Tuan and Rana, Santu and Gupta, Sunil and Venkatesh, Svetha},
    title     = {Learn To Unlearn for Deep Neural Networks: Minimizing Unlearning Interference With Gradient Projection},
    booktitle = {Proceedings of the IEEE/CVF Winter Conference on Applications of Computer Vision (WACV)},
    month     = {January},
    year      = {2024},
    pages     = {4819-4828}
}

@inproceedings{farajtabar2020orthogonal,
  title={Orthogonal gradient descent for continual learning},
  author={Farajtabar, Mehrdad and Azizan, Navid and Mott, Alex and Li, Ang},
  booktitle={International conference on artificial intelligence and statistics},
  pages={3762--3773},
  year={2020},
  organization={PMLR}
}

@article{sekhari2021remember,
  title={Remember what you want to forget: Algorithms for machine unlearning},
  author={Sekhari, Ayush and Acharya, Jayadev and Kamath, Gautam and Suresh, Ananda Theertha},
  journal={Advances in Neural Information Processing Systems},
  volume={34},
  pages={18075--18086},
  year={2021}
}

@inproceedings{eldan2023s,
  title={Revisiting who’s harry potter: Towards targeted unlearning from a causal intervention perspective},
  author={Liu, Yujian and Zhang, Yang and Jaakkola, Tommi and Chang, Shiyu},
  booktitle={Proceedings of the 2024 Conference on Empirical Methods in Natural Language Processing},
  pages={8708--8731},
  year={2024}
}

@article{liu2025rethinking,
  title={Rethinking machine unlearning for large language models},
  author={Liu, Sijia and Yao, Yuanshun and Jia, Jinghan and Casper, Stephen and Baracaldo, Nathalie and Hase, Peter and Yao, Yuguang and Liu, Chris Yuhao and Xu, Xiaojun and Li, Hang and others},
  journal={Nature Machine Intelligence},
  pages={1--14},
  year={2025},
  publisher={Nature Publishing Group UK London}
}

@article{meng2022locating,
  title={Locating and editing factual associations in gpt},
  author={Meng, Kevin and Bau, David and Andonian, Alex and Belinkov, Yonatan},
  journal={Advances in neural information processing systems},
  volume={35},
  pages={17359--17372},
  year={2022}
}

@article{belrose2023leace,
  title={Leace: Perfect linear concept erasure in closed form},
  author={Belrose, Nora and Schneider-Joseph, David and Ravfogel, Shauli and Cotterell, Ryan and Raff, Edward and Biderman, Stella},
  journal={Advances in Neural Information Processing Systems},
  volume={36},
  pages={66044--66063},
  year={2023}
}

@article{lynch2024eight,
  title={Eight methods to evaluate robust unlearning in llms},
  author={Lynch, Aengus and Guo, Phillip and Ewart, Aidan and Casper, Stephen and Hadfield-Menell, Dylan},
  journal={arXiv preprint arXiv:2402.16835},
  year={2024}
}

@article{zhang2024unlearncanvas,
  title={Unlearncanvas: A stylized image dataset to benchmark machine unlearning for diffusion models},
  author={Zhang, Yihua and Zhang, Yimeng and Yao, Yuguang and Jia, Jinghan and Liu, Jiancheng and Liu, Xiaoming and Liu, Sijia},
  journal={CoRR},
  year={2024}
}

@inproceedings{mehta2023esd,
  title={ESD: Efficient Style Disentanglement for Diffusion-based Image Unlearning},
  author={Mehta, Rohan and Dai, Zhijie and Gupta, Abhishek and Ghosh, Shankha and Wang, Xiaolong},
  booktitle={Proceedings of the IEEE/CVF International Conference on Computer Vision (ICCV)},
  year={2023}
}

@misc{krizhevsky2009learning,
  title={Learning multiple layers of features from tiny images.(2009)},
  author={Krizhevsky, Alex and Hinton, Geoffrey and others},
  year={2009}
}

@incollection{gholami2022survey,
  title={A survey of quantization methods for efficient neural network inference},
  author={Gholami, Amir and Kim, Sehoon and Dong, Zhen and Yao, Zhewei and Mahoney, Michael W and Keutzer, Kurt},
  booktitle={Low-power computer vision},
  pages={291--326},
  year={2022},
  publisher={Chapman and Hall/CRC}
}

@article{frantar2022gptq,
  title={Gptq: Accurate post-training quantization for generative pre-trained transformers},
  author={Frantar, Elias and Ashkboos, Saleh and Hoefler, Torsten and Alistarh, Dan},
  journal={arXiv preprint arXiv:2210.17323},
  year={2022}
}

@article{fan2023salun,
  title={Salun: Empowering machine unlearning via gradient-based weight saliency in both image classification and generation},
  author={Fan, Chongyu and Liu, Jiancheng and Zhang, Yihua and Wong, Eric and Wei, Dennis and Liu, Sijia},
  journal={arXiv preprint arXiv:2310.12508},
  year={2023}
}

@article{xie2025ec,
  title={EC-Diff: Fast and High-Quality Edge-Cloud Collaborative Inference for Diffusion Models},
  author={Xie, Jiajian and Zhang, Shengyu and Zhao, Zhou and Wu, Fan and Wu, Fei},
  journal={arXiv preprint arXiv:2507.11980},
  year={2025}
}

@inproceedings{gandikota2024unified,
  title={Unified concept editing in diffusion models},
  author={Gandikota, Rohit and Orgad, Hadas and Belinkov, Yonatan and Materzy{\'n}ska, Joanna and Bau, David},
  booktitle={Proceedings of the IEEE/CVF Winter Conference on Applications of Computer Vision},
  pages={5111--5120},
  year={2024}
}

@article{fu2023mme,
  title={MME: A Comprehensive Evaluation Benchmark for Multimodal Large Language Models},
  author={Fu, Chaoyi and Xu, Haotian and Wu, Yixuan and others},
  journal={arXiv preprint arXiv:2306.13394},
  year={2023}
}

@inproceedings{hudson2019gqa,
  title={GQA: A New Dataset for Real-World Visual Reasoning and Compositional Question Answering},
  author={Hudson, Drew A. and Manning, Christopher D.},
  booktitle={Proceedings of the IEEE/CVF Conference on Computer Vision and Pattern Recognition (CVPR)},
  year={2019}
}

@article{liu2025mmbench,
  title={MMBench: Comprehensive Multimodal Evaluation Benchmark for Vision–Language Models},
  author={Liu, Haotian and Li, Chunyuan and Li, Yuheng and Lee, Yong Jae},
  journal={arXiv preprint arXiv:2501.03145},
  year={2025}
}

@article{zhang2024catastrophic,
  title={Catastrophic failure of llm unlearning via quantization},
  author={Zhang, Zhiwei and Wang, Fali and Li, Xiaomin and Wu, Zongyu and Tang, Xianfeng and Liu, Hui and He, Qi and Yin, Wenpeng and Wang, Suhang},
  journal={arXiv preprint arXiv:2410.16454},
  year={2024}
}

@article{siddiqui2025from,
  title   = {From Dormant to Deleted: Tamper-Resistant Unlearning Through Weight-Space Regularization},
  author  = {Siddiqui, Shoaib Ahmed and Weller, Adrian and Krueger, David and Dziugaite, Gintare Karolina and Mozer, Michael Curtis and Triantafillou, Eleni},
  journal = {arXiv preprint arXiv:2505.22310},
  year    = {2025}
}

@article{liu2023visual,
  title={Visual instruction tuning},
  author={Liu, Haotian and Li, Chunyuan and Wu, Qingyang and Lee, Yong Jae},
  journal={Advances in neural information processing systems},
  volume={36},
  pages={34892--34916},
  year={2023}
}

@article{mistretta2025cross,
  title={Cross the gap: Exposing the intra-modal misalignment in clip via modality inversion},
  author={Mistretta, Marco and Baldrati, Alberto and Agnolucci, Lorenzo and Bertini, Marco and Bagdanov, Andrew D},
  journal={arXiv preprint arXiv:2502.04263},
  year={2025}
}

@inproceedings{cong2025guiding,
  title={Guiding noisy label conditional diffusion models with score-based discriminator correction},
  author={Cong, Dat Nguyen and Bao, Hieu Tran and Hoang-Thanh, Tung},
  booktitle={Proceedings of the IEEE/CVF International Conference on Computer Vision},
  pages={18531--18541},
  year={2025}
}

@article{wang2024understanding,
  title={Understanding and mitigating miscalibration in prompt tuning for vision-language models},
  author={Wang, Shuoyuan and Li, Yixuan and Wei, Hongxin},
  journal={arXiv preprint arXiv:2410.02681},
  year={2024}
}

@inproceedings{kumari2023ablating,
  title={Ablating concepts in text-to-image diffusion models},
  author={Kumari, Nupur and Zhang, Bingliang and Wang, Sheng-Yu and Shechtman, Eli and Zhang, Richard and Zhu, Jun-Yan},
  booktitle={Proceedings of the IEEE/CVF International Conference on Computer Vision},
  pages={22691--22702},
  year={2023}
}

@misc{ranjan2026positionllmsusefunctorbased,
      title={Position: LLMs Must Use Functor-Based and RAG-Driven Bias Mitigation for Fairness}, 
      author={Ravi Ranjan and Utkarsh Grover and Agorista Polyzou},
      year={2026},
      eprint={2603.07368},
      archivePrefix={arXiv},
      primaryClass={cs.CL},
      url={https://arxiv.org/abs/2603.07368}, 
}

@article{ranjan2026position,
  title={Position: LLMs Must Use Functor-Based and RAG-Driven Bias Mitigation for Fairness},
  author={Ranjan, Ravi and Grover, Utkarsh and Polyzou, Agorista},
  journal={arXiv preprint arXiv:2603.07368},
  year={2026}
}
}


\clearpage
\appendix
\onecolumn

\section{Pseudocode for RAZOR}
\label{app:pseudo}
\newtcolorbox{razoralgobox}[1][]{
  colback=blue!3,
  colframe=blue!60!black,
  coltitle=white,
  fonttitle=\bfseries,
  title=#1,
  breakable,
  enhanced,
  boxrule=0.8pt,
  arc=2pt,
  left=4pt,
  right=4pt,
  top=6pt,
  bottom=6pt
}

\SetAlFnt{\small}
\SetKwInput{Input}{Input}
\SetKwInput{Output}{Output}
\SetKwFunction{BinarySearchStep}{BinarySearchStep}
\SetKwFunction{EvaluateMetrics}{EvaluateMetrics}
\SetKwFunction{ScoreMetrics}{ScoreMetrics}

\begin{razoralgobox}[RAZOR: Ratio-Aware Layer Editing for Targeted Unlearning]
\begin{algorithm}[H]
\DontPrintSemicolon

\Input{
Pretrained model $f_\theta$ with parameters $\theta$ and components $L = \{1,\dots,|L|\}$ \\
Forget set $D_f$, retain set $D_r$, validation split $D_{\text{val}}$ \\
Ratio hyperparameter $\rho$; coefficients $\lambda_f, \lambda_m$ \\
Orthogonality exponent $\alpha$; stability constant $\epsilon$ \\
Initial saliency threshold $\tau_{\text{init}}$; max iterations $T_{\max}$ \\
Metric constraints $\text{Target}$ (e.g., thresholds on M1--M5)
}
\Output{Edited parameters $\theta^\ast$}

\BlankLine
\textbf{Stage 0: Baseline statistics for mismatch loss}\;
$\theta^{0} \leftarrow \theta$ \tcp*{Frozen copy of initial model}
Compute baseline signals (e.g., embeddings, logits) for $L_{\text{mismatch}}$ using $f_{\theta^{0}}$ on $D_f$\;

\BlankLine
\textbf{Stage 1: One-shot gradients \& ratio-aware saliency}\;
Compute $g_f^\ell = \nabla_{\theta_\ell} L_{\text{forget}}(\theta; D_f)$ for all $\ell \in L$\;
Compute $g_r^\ell = \nabla_{\theta_\ell} L_{\text{retain}}(\theta; D_r)$ for all $\ell \in L$\;

\ForEach{$\ell \in L$}{
  $\text{num} \leftarrow \| g_f^\ell \|_2^2$ , 
  $\text{denom} \leftarrow \| \theta_\ell \|_2^2 + \epsilon$\;
  $\phi(\ell) \leftarrow \left(\dfrac{\text{num}}{\text{denom}}\right)\bigl(1 - \cos(g_f^\ell, g_r^\ell)\bigr)^\alpha$\;
}

$\mathcal{K} \leftarrow \{\,\ell : \phi(\ell) > \tau_{\text{init}}\,\}$\;
\If{$\mathcal{K} = \emptyset$}{
  $\mathcal{K} \leftarrow \{\arg\max_{\ell} \phi(\ell)\}$ \tcp*{Ensure at least one component}
}

\BlankLine
\textbf{Stage 2: Update initially selected components}\;
\ForEach{$\ell \in \mathcal{K}$}{
  Compute $g_m^\ell = \nabla_{\theta_\ell} L_{\text{mismatch}}(\theta; D_f, \theta^{0})$\;
  $g_{\text{RAZOR}}^\ell \leftarrow -\lambda_f \rho\, g_f^\ell + g_r^\ell + \lambda_m g_m^\ell$\;
  $\lambda_\ell \leftarrow \BinarySearchStep(\theta, \ell, g_{\text{RAZOR}}^\ell, D_{\text{val}}, \text{Target})$\;
  $\theta_\ell \leftarrow \theta_\ell - \lambda_\ell g_{\text{RAZOR}}^\ell$\;
}

\BlankLine
\textbf{Stage 3: Iterative refinement of the active set}\;
$t \leftarrow 1$\;
\While{$t \leq T_{\max}$}{
  $\text{metrics} \leftarrow \EvaluateMetrics(f_\theta, D_{\text{val}})$\;
  \If{$\text{metrics}$ satisfy $\text{Target}$}{
    \textbf{break}\tcp*{Desired forgetting/utility achieved}
  }

  Recompute $g_f^\ell$ and $g_r^\ell$ for all $\ell \in L$\;

  \ForEach{$\ell \in L$}{
    $\text{num} \leftarrow \| g_f^\ell \|_2^2$ ,
    $\text{denom} \leftarrow \| \theta_\ell \|_2^2 + \epsilon$\;
    $\phi(\ell) \leftarrow \left(\dfrac{\text{num}}{\text{denom}}\right)\bigl(1 - \cos(g_f^\ell, g_r^\ell)\bigr)^\alpha$\;
  }

  $\ell^\ast \leftarrow \arg\max_{\ell \notin K} \phi(\ell)$\;
  \If{$\phi(\ell^\ast) \le 0$}{
    \textbf{break}\tcp*{No useful additional component to edit}
  }

  $\mathcal{K} \leftarrow \mathcal{K} \cup \{\ell^\ast\}$\;

  Compute $g_m^{\ell^\ast}$ and $g_{\text{RAZOR}}^{\ell^\ast}$ as above\;
  $\lambda_{\ell^\ast} \leftarrow \BinarySearchStep(\theta, \ell^\ast, g_{\text{RAZOR}}^{\ell^\ast}, D_{\text{val}}, \text{Target})$\;
  $\theta_{\ell^\ast} \leftarrow \theta_{\ell^\ast} - \lambda_{\ell^\ast} g_{\text{RAZOR}}^{\ell^\ast}$\;

  $t \leftarrow t + 1$\;
}
\Return{$\theta^\ast \leftarrow \theta$}\;

\caption{RAZOR: Ratio-Aware Layer Editing for Targeted Unlearning}
\label{alg:razor}
\end{algorithm}
\end{razoralgobox}

\begin{razoralgobox}[Binary search for per-layer step size $\lambda_l$]
\begin{algorithm}[H]
\DontPrintSemicolon

\Input{
Current parameters $\theta$; selected component index $l$; blended gradient $g_{\text{RAZOR}}^{\,l}$ \\
Validation set $D_{\text{val}}$; target metric constraints $\textsc{Target}$
}
\Output{Layer-local step size $\lambda_l$}

$\lambda_{\min} \leftarrow 0$, \quad $\lambda_{\max} \leftarrow \lambda_{\text{init}}$\;
$\lambda_{\text{best}} \leftarrow 0$, \quad $s_{\text{best}} \leftarrow -\infty$\;

\While{$\lambda_{\max} - \lambda_{\min} > \delta$}{
  $\lambda_{\text{mid}} \leftarrow (\lambda_{\min} + \lambda_{\max}) / 2$\;

  \tcp{Propose a temporary RAZOR update on component $l$}
  $\theta^{\text{temp}} \leftarrow \theta$\;
  $\theta^{\text{temp}}_l \leftarrow \theta_l - \lambda_{\text{mid}}\, g_{\text{RAZOR}}^{\,l}$\;

  \tcp{Evaluate forgetting/retention trade-off under the proposal}
  $\text{metrics} \leftarrow \EvaluateMetrics(f_{\theta^{\text{temp}}}, D_{\text{val}})$\;
  $s \leftarrow \ScoreMetrics(\text{metrics}, \textsc{Target})$\;

  \If{\text{model is stable and meets basic constraints under $\theta^{\text{temp}}$}}{
    \If{$s > s_{\text{best}}$}{
      $s_{\text{best}} \leftarrow s$\;
      $\lambda_{\text{best}} \leftarrow \lambda_{\text{mid}}$\;
    }
    $\lambda_{\min} \leftarrow \lambda_{\text{mid}}$ \tcp*{Safe to try a larger step}
  }
  \Else{
    $\lambda_{\max} \leftarrow \lambda_{\text{mid}}$ \tcp*{Step too large; shrink the interval}
  }
}

\Return{$\lambda_{\text{best}}$}\;

\label{alg:razor-binary-search}
\end{algorithm}
\end{razoralgobox}

\section{Detailed Related Work}
\label{app:related}
This section is an expanded version of the Related Work in Section \ref{sec:related-work}, offering deeper technical context, additional citations, and broader coverage of prior unlearning methods across CLIP, diffusion, and VLM architectures.

\noindent Machine unlearning \citep{cao2015towards, nguyen2022survey,foster2024fast,shaik2024exploring,bourtoule2021machine} addresses the selective removal of data influence from trained models, a critical need driven by privacy concerns and regulatory requirements \citep{carmi2023european}. Existing approaches mainly focus on a single task, like image classification \cite{jia2023model, triantafillou2023evaluation,  goel2022towards, wu2023certified, shah2023unlearning,  kurmanji2023towards}, image generation \cite{gandikota2023erasing, zhang2024unlearncanvas,wu2024scissorhands, kurmanji2023towards, wu2023certified,zhang2024forget, wu2025erasing,lyu2024one}, and LLMs text generation \cite{yao2024large}. In this work, we propose a generic approach that is applicable to a wide range of multimodal models, including \textsc{CLIP} \citep{radford2021clip} for zero-shot image classification, stable diffusion models \citep{rombach2022high} for text-to-image generation, and vision language models \citep{liu2024improved} for visual question answering.

\noindent \textbf{Exact vs. approximate unlearning.} The gold standard, exact removal (retraining on retain set), offers provable guarantees \citep{guo2023certifieddataremovalmachine} but is computationally infeasible for large scale models \citep{cottier2024rising, guerra2023cost}. Research has thus shifted to \emph{approximate unlearning} methods that modify parameters \citep{cao2015towards, ginart2019making}. SISA \citep{bourtoule2021machine}, a middle ground, trains on data shards for efficient partial retraining but requires architectural foresight and incurs storage overhead. We focus on \emph{post hoc unlearning}: editing pre trained models without access to the original training pipeline \citep{nguyen2025survey,zhang2023review,wang2023machine}.
Post hoc unlearning must identify which parameters to modify, as naïve full model updates risk catastrophic forgetting \citep{wang2023machine}. While early work used imprecise uniform perturbations \citep{golatkar2020forgetting}, gradient-based saliency has emerged as the dominant localization paradigm. \textsc{SalUn} \citep{fan2024salun} ranks parameters and fine tunes only high saliency weights. \textsc{Scissorhands} \citep{wu2024scissorhands} improves this by computing gradients at initialization to reinitialize connections before recovery.
Higher order methods, like Fisher-based Selective Synaptic Dampening \citep{foster2024loss}, use curvature to preserve important weights, but computing the Fisher matrix scales poorly \citep{martens2020new}. As a compromise, \textsc{SLUG} \citep{cai2025slug} trades completeness for efficiency by updating only a \emph{single} high-impact layer. The other argument is that we can modify LLM behavior using mathematics~\cite{ranjan2026positionllmsusefunctorbased, ranjan2026position}. These approaches highlight the core tension: localized edits are efficient but brittle, while exhaustive updates are robust but prohibitive.

\noindent Given \emph{which parameters}, the challenge becomes \emph{how to update them}, navigating the conflicting gradients from retain set and forget set \citep{yu2020gradient}. Direct gradient ascent on the  forget set \citep{thudi2022unrolling} often destroys generalization. Gradient projection techniques \citep{Hoang_2024_WACV,patel2025learning} mitigate this by constraining updates to subspaces orthogonal to the retained loss, preserving retain set performance, similar to methods in continual learning \citep{farajtabar2020orthogonal}. Representation space methods offer an alternative. Contrastive Unlearning repels forget set embeddings from their class centroids \citep{kyu2024contrastive}. Bad Teaching distills knowledge away from the forget set \citep{liu2024improved}. However, these parameter-level methods require full backpropagation and are unstable when the forget and retain sets have overlapping support. \textsc{LoTUS} smooths forget set predictions toward a uniform distribution \citep{spartalis2025lotus}. This handles concept-level unlearning but struggles with instance-level deletion and often requires forget set labels \citep{sekhari2021remember}.

\noindent \textbf{Unlearning in Generative Models.} These discriminative model approaches are ill-suited for generative models, where concepts are distributed compositionally \citep{rombach2022high}, rendering classifier based localization ineffective \citep{gandikota2023erasing}. Inference time interventions like Erased Stable Diffusion (ESD) \citep{gandikota2023erasing} apply negative guidance, but this only \emph{suppresses} expression, leaving knowledge recoverable via adversarial prompts \citep{zhang2024unlearncanvas}. This limitation motivates activation level surgery, such as using Sparse Autoencoders (SAEs) to find and ablate interpretable features (\textsc{SAeUron} \citep{cywinski2025saeuron}) or dynamically masking gradients (\citet{fan2024salun}). LLM unlearning faces analogous issues \citep{eldan2023s, liu2025rethinking, meng2022locating, belrose2023leace}, as creative prompting can often recover ``forgotten'' information \citep{lynch2024eight}.

\noindent Existing approaches reveal fundamental limitations that constrain practical unlearning. First, localization retention trade-offs remain unresolved: methods like \textsc{SalUn} \citep{fan2024salun} and \textsc{Scissorhands} \citep{wu2024scissorhands} identify salient parameters through gradient magnitude but lack explicit mechanisms to balance forgetting pressure against retention requirements, often leading to overly aggressive edits that degrade utility. Conversely, ultra conservative approaches, like \textsc{SLUG} \citep{cai2025slug}, which confine edits to a single layer, achieve strong retention but struggle when knowledge is distributed across multiple architectural components. Second, gradient conflict management remains ad hoc: while projection methods \citep{patel2025learning} and contrastive techniques \citep{kyu2024contrastive} mitigate opposing gradients, they do so through post-hoc constraints rather than jointly optimizing for the forget-retain trade-off during parameter selection. This reactive approach cannot prevent conflicts that arise from poor localization choices. Third, architectural generalization is limited: most methods are designed for specific classifier model families, diffusion models, or language models, and require substantial re-engineering to transfer across domains. Methods effective for discriminative models often fail for generative architectures due to their fundamentally different knowledge encoding mechanisms \citep{gandikota2023erasing, xiao2025reminiscence}.

\noindent We introduce \textbf{RAZOR}, a ratio-aware framework that unifies parameter localization and update computation. RAZOR scores layers and heads by a forget-to-retain gradient ratio, quantifying the forget-pressure vs. retention-alignment trade-off. This enables principled multi-layer selection, avoiding both the brittleness of single-layer edits (e.g., \textsc{SLUG}) and the inefficiency of exhaustive updates. Its objective composes three losses (retain, forget, mismatch) governed by a ratio $\rho$ to preemptively resolve gradient conflicts. RAZOR's formulation generalizes across vision encoders (ViT, \textsc{CLIP}), diffusion text encoders (Stable Diffusion), and vision-language models (\textsc{LLaVA}), requiring only loss instantiation, not architectural redesign. Through iterative layer-wise refinement, it matches the precision of methods like Selective Synaptic Dampening with the efficiency of \textsc{SLUG} (updating only $k \ll L$ components), thus reconciling the efficiency-effectiveness-robustness trilemma.

\section{Propositions Supporting RAZOR}
\label{app:theory}

\textbf{Proposition 1 (Convergence).}
\textit{The RAZOR layer-editing process converges to a minimal set of edited layers $K$ achieving the desired forgetting criterion.}
RAZOR begins with a saliency-thresholded subset $K \subseteq L$ based on ratio-aware scores $\varphi(l)$. If this edit is insufficient, RAZOR greedily expands $K$ by appending the next most salient layer and applying a targeted update. Since each layer edit maximizes forgetting relative to retention loss, and there are finitely many layers, the sequence $K_0 \subset K_1 \subset \cdots$ converges in at most $|L|$ iterations (typically $\le6$ in practice). Crucially, per-layer updates use binary search to find the largest stable step size, ensuring convergence without catastrophic drift.

\medskip
\textbf{Proposition 2 (Forgetting Guarantee).}
\textit{The combined forgetting and mismatch losses enforce representation drift and suppress alignment on the forget set.}
RAZOR minimizes $L_{\text{RAZOR}} = L_{\text{retain}} + \lambda_f \rho L_{\text{forget}} + \lambda_m L_{\text{mismatch}}$, where $L_{\text{forget}}$ reduces image–text similarity on $D_f$, and $L_{\text{mismatch}}$ penalizes similarity to the frozen base model. Together, these ensure that for any $(v,t)\in D_f$, the final representation $\langle v,t \rangle$ is actively suppressed below both its original value and the retention threshold. At optimality, target knowledge is provably erased from the model’s embedding space.

\medskip
\textbf{Proposition 3 (Retention–Forgetting Trade-off).}
\textit{RAZOR imposes a bounded utility loss, controlled by the ratio $\rho \in (0,1]$.}
The term $\rho$ functions as a Lagrange multiplier that trades forgetting accuracy for retention fidelity. At convergence, we obtain a Pareto-optimal solution balancing $\nabla_\theta L_{\text{retain}}$ and $\nabla_\theta L_{\text{forget}}$, such that retention degradation scales linearly with $\rho$. Thus, $\Delta L_{\text{retain}} \lesssim \rho \cdot \Delta L_{\text{forget}}$, providing a tunable bound on performance drop. This ensures RAZOR avoids under-forgetting (like SalUn) or over-forgetting (like naive fine-tuning).

\medskip
\textbf{Proposition 4 (Saliency Justification).}
\textit{RAZOR’s saliency score $\varphi(l)$ identifies high-impact, low-interference edits.}
Each component’s score is defined as
\[
\varphi(l) = \frac{\|\nabla_{\theta_l} L_{\text{forget}}\|_2}{\|\theta_l\|_2 + \epsilon} \cdot (1 - \cos(\nabla_{\theta_l} L_{\text{forget}}, \nabla_{\theta_l} L_{\text{retain}}))^\alpha
\]
where $\alpha \in [0,1]$. The first term captures forgetting impact (gradient norm), while the second downweights layers where forgetting and retention gradients align. As such, $\varphi(l)$ is large only when forgetting can be achieved orthogonally to retention directions, minimizing interference. This principled selection improves over prior magnitude-only (e.g., Fisher) or shallow metrics.

\medskip
Together, these four propositions establish that RAZOR converges efficiently, provably forgets target knowledge, maintains bounded utility degradation, and performs theoretically justified low-interference edits.

\section{RAZOR Losses}
\label{app:experimental}

In this section we will describe the Stable diffusion and VLM loss functions mentioned in table-\ref{tab:meth-1}, section \ref{sec:method:instantiation}.
\subsection{RAZOR Losses for Stable Diffusion}
For text-to-image diffusion models, we instantiate the RAZOR objective
\(
\mathcal{L}_{\text{RAZOR}}
= \mathcal{L}_{\text{retain}}
+ \lambda_f \rho\, \mathcal{L}_{\text{forget}}
+ \lambda_m \mathcal{L}_{\text{mismatch}}
\)
using losses that operate on the text encoder and guidance scores of the UNet, following the entries reported for Stable Diffusion in Table~\ref{sec:method:instantiation}. 

\paragraph{(a) Retain loss $\mathcal{L}_{\text{retain}}^{\text{SD}}$.}
To preserve generative quality and alignment on prompts that must remain valid, we adopt the standard $\epsilon$-prediction denoising objective used to train diffusion models~\cite{rombach2022high}.  
Let $(x_i, t_i) \in \mathcal{D}_r$ be retain images and prompts, $e_i = f_t(t_i)$ the corresponding text embeddings, and 
\(
x_t = \sqrt{\alpha_t} x_i + \sqrt{1-\alpha_t}\,\varepsilon
\)
the noisy latent at time step $t$ with $\varepsilon \sim \mathcal{N}(0,\mathbf{I})$.  
The UNet predicts $\varepsilon_\theta(x_t, t, e_i)$, and the retain loss is
\begin{equation}
\mathcal{L}_{\text{retain}}^{\text{SD}}(\theta; \mathcal{D}_r)
=
\frac{1}{|\mathcal{D}_r|}
\sum_{(x_i, t_i) \in \mathcal{D}_r}
\mathbb{E}_{t,\,\varepsilon}
\Big[
\big\|
\varepsilon - \varepsilon_\theta(x_t, t, e_i)
\big\|_2^2
\Big],
\label{eq:sd_retain}
\end{equation}
which encourages the edited model to match the original denoising behavior on retain prompts.

\paragraph{(b) Forget loss $\mathcal{L}_{\text{forget}}^{\text{SD}}$.}
For Stable Diffusion, forgetting is driven at the level of the text encoder $f_t$, using a cross-entropy objective on forget prompts~\cite{cai2025slug}.  
Let $\mathcal{D}_f$ be the forget prompt set, $e_i = f_t(t_i)$ their embeddings, and $z_\theta(e_i)$ the logits of a small classification head (e.g., “forget concept’’ vs. “other’’).  
Denote by $p_\theta(c \mid t_i)$ the induced probabilities and by $q_i(c)$ a target distribution that down-weights the forget concept (e.g., mass moved to a neutral or ``other'' class).  
We define
\begin{equation}
\mathcal{L}_{\text{forget}}^{\text{SD}}(\theta; \mathcal{D}_f)
=
-\frac{1}{|\mathcal{D}_f|}
\sum_{t_i \in \mathcal{D}_f}
\sum_{c}
q_i(c)\,
\log p_\theta(c \mid t_i),
\label{eq:sd_forget}
\end{equation}
which explicitly suppresses the association between forget prompts and their original concept labels.

\paragraph{(c) Mismatch loss $\mathcal{L}_{\text{mismatch}}^{\text{SD}}$.}
To control drift in the generative behavior and stabilize guidance, we employ a Similarity Drift Regularizer (SDR) on guidance/similarity scores for generated samples~\cite{kumari2023ablating, cong2025guiding}.  
Let $\theta^{(0)}$ be the frozen pre-edit parameters, and let $s_\theta(x_t, t, e)$ denote a scalar guidance or similarity score used during sampling (e.g., classifier-free guidance score, CLIP-based alignment, or a logit used for guidance).  
We evaluate drift on a pool $\mathcal{G}$ of generated trajectories obtained from both forget and retain prompts:
\begin{equation}
\mathcal{L}_{\text{mismatch}}^{\text{SD}}(\theta; \mathcal{G})
=
\frac{1}{|\mathcal{G}|}
\sum_{(x_t, t, e) \in \mathcal{G}}
\big(
s_\theta(x_t, t, e)
-
s_{\theta^{(0)}}(x_t, t, e)
\big)^2.
\label{eq:sd_mismatch}
\end{equation}
This term anchors the edited model’s guidance behavior to the original model, preventing excessive changes in similarity structure while the forget loss pushes the targeted concepts away.

\subsection{RAZOR Losses for Vision Language Models (LLaVA)}

For vision language models such as LLaVA-1.6, RAZOR operates on the vision encoder and its alignment with textual concepts, while respecting the downstream multimodal behavior of the LLM head. We instantiate
\[
\mathcal{L}_{\text{RAZOR}}
=
\mathcal{L}_{\text{retain}}
+ \lambda_f \rho\, \mathcal{L}_{\text{forget}}
+ \lambda_m \mathcal{L}_{\text{mismatch}}
\]
using the VLM-specific losses summarized for LLaVA in Table~1.

\paragraph{(a) Retain loss $\mathcal{L}_{\text{retain}}^{\text{VLM}}$.}
To preserve visual utility and concept alignment for retained identities and objects, we adopt a symmetric InfoNCE contrastive loss on the \emph{vision encoder} tokens, following the visual instruction tuning setup of LLaVA~\cite{liu2024improved, liu2023visual}.  
Let $(x_i, y_i) \in \mathcal{D}_r$ be retain images and their textual concept prompts (or captions), and let $v_i = f_v(x_i)$ denote pooled visual embeddings, while $t_i = f_t(y_i)$ are text embeddings obtained from the text/LLM encoder (or a frozen text tower). We define
\begin{equation}
\mathcal{L}_{\text{retain}}^{\text{VLM}}(\theta; \mathcal{D}_r)
=
\frac{1}{2|\mathcal{D}_r|}
\sum_{i}
\Bigg[
- \log
\frac{\exp(\langle v_i, t_i\rangle / \tau)}{\sum_j \exp(\langle v_i, t_j\rangle / \tau)}
-
\log
\frac{\exp(\langle v_i, t_i\rangle / \tau)}{\sum_j \exp(\langle v_j, t_i\rangle / \tau)}
\Bigg],
\label{eq:vlm_retain}
\end{equation}
where $\tau > 0$ is a temperature. This term encourages the edited vision encoder to maintain strong alignment between retain images and their textual descriptions.

\paragraph{(b) Forget loss $\mathcal{L}_{\text{forget}}^{\text{VLM}}$.}
For identity and concept unlearning in VLMs, we follow the CLIP-style formulation and apply a cross-entropy ``push-away'' loss on the \emph{vision encoder} for forget concepts~\cite{cai2025slug}.  
Let $\mathcal{D}_f$ be the forget set consisting of images containing a target concept (e.g., a specific celebrity), with visual embeddings $v_i = f_v(x_i)$ and a small concept classifier head with logits $z_\theta(v_i) \in \mathbb{R}^C$. We construct target distributions $q_i(c)$ that down-weight the forgotten concept $c_f$ (e.g., reassigning its probability mass to a neutral or ``other'' class). The forget loss is
\begin{equation}
\mathcal{L}_{\text{forget}}^{\text{VLM}}(\theta; \mathcal{D}_f)
=
- \frac{1}{|\mathcal{D}_f|}
\sum_{x_i \in \mathcal{D}_f}
\sum_{c=1}^C
q_i(c)\,
\log p_\theta(c \mid v_i),
\label{eq:vlm_forget}
\end{equation}
where $p_\theta(c \mid v_i)$ is the softmax probability from $z_\theta(v_i)$. This term explicitly suppresses the visual encoding of the forgotten identity or object at the concept level.

\paragraph{(c) Mismatch loss $\mathcal{L}_{\text{mismatch}}^{\text{VLM}}$.}
To regularize the multimodal behavior and prevent unintended drift on neutral tasks, we employ a similarity/logit-drift regularizer on neutral QA prompts and captions~\cite{wang2024understanding}.  
Let $\theta^{(0)}$ denote the frozen base model, and consider a set $\mathcal{N}$ of neutral image--question pairs $(x_i, q_i)$ that do not mention forgotten concepts. For each pair, the VLM produces pre-softmax logits $h_\theta(x_i, q_i)$ (e.g., over answer tokens or a pooled scoring head) and corresponding base logits $h_{\theta^{(0)}}(x_i, q_i)$. We define
\begin{equation}
\mathcal{L}_{\text{mismatch}}^{\text{VLM}}(\theta; \mathcal{N})
=
\frac{1}{|\mathcal{N}|}
\sum_{(x_i, q_i) \in \mathcal{N}}
\big\|
h_\theta(x_i, q_i)
-
h_{\theta^{(0)}}(x_i, q_i)
\big\|_2^2,
\label{eq:vlm_mismatch}
\end{equation}
which penalizes large deviations in the VLM’s logits on neutral inputs.  
This term stabilizes the unlearning procedure by anchoring the edited model to the base model on non-forget queries, while $\mathcal{L}_{\text{forget}}^{\text{VLM}}$ and $\mathcal{L}_{\text{retain}}^{\text{VLM}}$ drive targeted erasure and utility preservation, respectively.

\begin{table}[b]
\centering
\scriptsize
\caption{Per–identity comparison of RAZOR and SLUG on LLaVA-1.6-8B for CelebA
identity unlearning. FA (↓) is forget accuracy; higher is better for all
other metrics.}
\label{tab:razor-vs-slug-llava-identity}
\setlength{\tabcolsep}{2.5pt}
\begin{tabular}{lrrrrrrrrrr}
\toprule
& \multicolumn{2}{c}{FA $\downarrow$}
& \multicolumn{2}{c}{MME (Cogn.) $\uparrow$}
& \multicolumn{2}{c}{MME (Perc.) $\uparrow$}
& \multicolumn{2}{c}{GQA $\uparrow$}
& \multicolumn{2}{c}{MMBench $\uparrow$} \\
\cmidrule(lr){2-3}
\cmidrule(lr){4-5}
\cmidrule(lr){6-7}
\cmidrule(lr){8-9}
\cmidrule(lr){10-11}
\textbf{Identity} &
\textbf{RAZOR} & \textbf{SLUG} &
\textbf{RAZOR} & \textbf{SLUG} &
\textbf{RAZOR} & \textbf{SLUG} &
\textbf{RAZOR} & \textbf{SLUG} &
\textbf{RAZOR} & \textbf{SLUG} \\
\midrule
Scarlett Johansson
& 2.0 & 3.0
& 301.4 & 301.6
& 1362.7 & 1365.5
& 60.83 & 60.40
& 61.90 & 61.86 \\

Taylor Swift
& 1.5 & 2.0
& 339.3 & 334.6
& 1347.9 & 1336.1
& 60.86 & 60.72
& 60.77 & 60.14 \\

Robert Downey Jr.
& 5.0 & 3.0
& 348.3 & 341.8
& 1217.2 & 1225.6
& 58.85 & 59.40
& 56.89 & 55.58 \\

Jeff Bezos
& 2.0 & 3.0
& 319.9 & 314.6
& 1328.7 & 1315.3
& 60.55 & 60.40
& 61.72 & 61.43 \\

Kanye West
& 3.0 & 4.0
& 319.1 & 314.6
& 1378.4 & 1365.5
& 61.36 & 61.17
& 61.90 & 61.68 \\

Tom Cruise
& 0.0 & 0.0
& 354.8 & 351.8
& 1428.6 & 1413.0
& 61.25 & 61.13
& 62.14 & 61.86 \\

Kim Kardashian
& 4.0 & 6.0
& 292.7 & 286.4
& 1281.8 & 1249.5
& 60.57 & 60.42
& 60.72 & 60.14 \\

Barack Obama
& 0.0 & 0.0
& 293.6 & 288.6
& 1302.5 & 1269.5
& 60.82 & 60.68
& 61.44 & 61.08 \\

Lady Gaga
& 2.0 & 3.0
& 274.9 & 270.4
& 1222.4 & 1178.5
& 58.85 & 58.55
& 56.20 & 55.58 \\

Natalie Portman
& 0.0 & 2.0
& 328.7 & 292.6
& 1315.2 & 1314.3
& 60.63 & 60.40
& 60.52 & 58.85 \\
\midrule
\textbf{Average}
& $1.9 \pm 1.6$ & $2.6 \pm 1.7$
& $317.3 \pm 24.9$ & $309.7 \pm 25.2$
& $1318.5 \pm 63.0$ & $1303.3 \pm 68.4$
& $60.5 \pm 0.8$ & $60.3 \pm 0.8$
& $60.4 \pm 2.0$ & $59.8 \pm 2.3$ \\
\bottomrule
\end{tabular}
\end{table}

\section{Result Comparison}
\label{app:result}

Continuing from Table~\ref{tab:razor_llava}, we compare SLUG with RAZOR. The results in Table~\ref{tab:razor-vs-slug-llava-identity} show that RAZOR consistently outperforms SLUG on identity unlearning for LLaVA-1.6-8B across all evaluation metrics. In terms of forgetting, RAZOR achieves a lower mean FA ($1.9 \pm 1.6$) compared to SLUG ($2.6 \pm 1.7$), indicating more effective suppression of the targeted identities while maintaining comparable variance. At the same time, RAZOR slightly improves utility-oriented scores, with higher averages on both MME (Cognition and Perception), GQA, and MMBench, and similar or smaller standard deviations. Importantly, these gains are not driven by a few outliers: for almost every identity, RAZOR attains either equal or better FA while matching or exceeding SLUG on downstream benchmarks. Overall, the table highlights that ratio-aware multi-layer editing provides a more favorable forget–retain trade-off than single-layer updates, enabling stronger identity removal without sacrificing multi-modal reasoning performance.

\section{Effect of Learning Rate on Performance}
\label{app:ablation}

\begin{figure}[hb]
   \centering\includegraphics[width=0.9\linewidth]{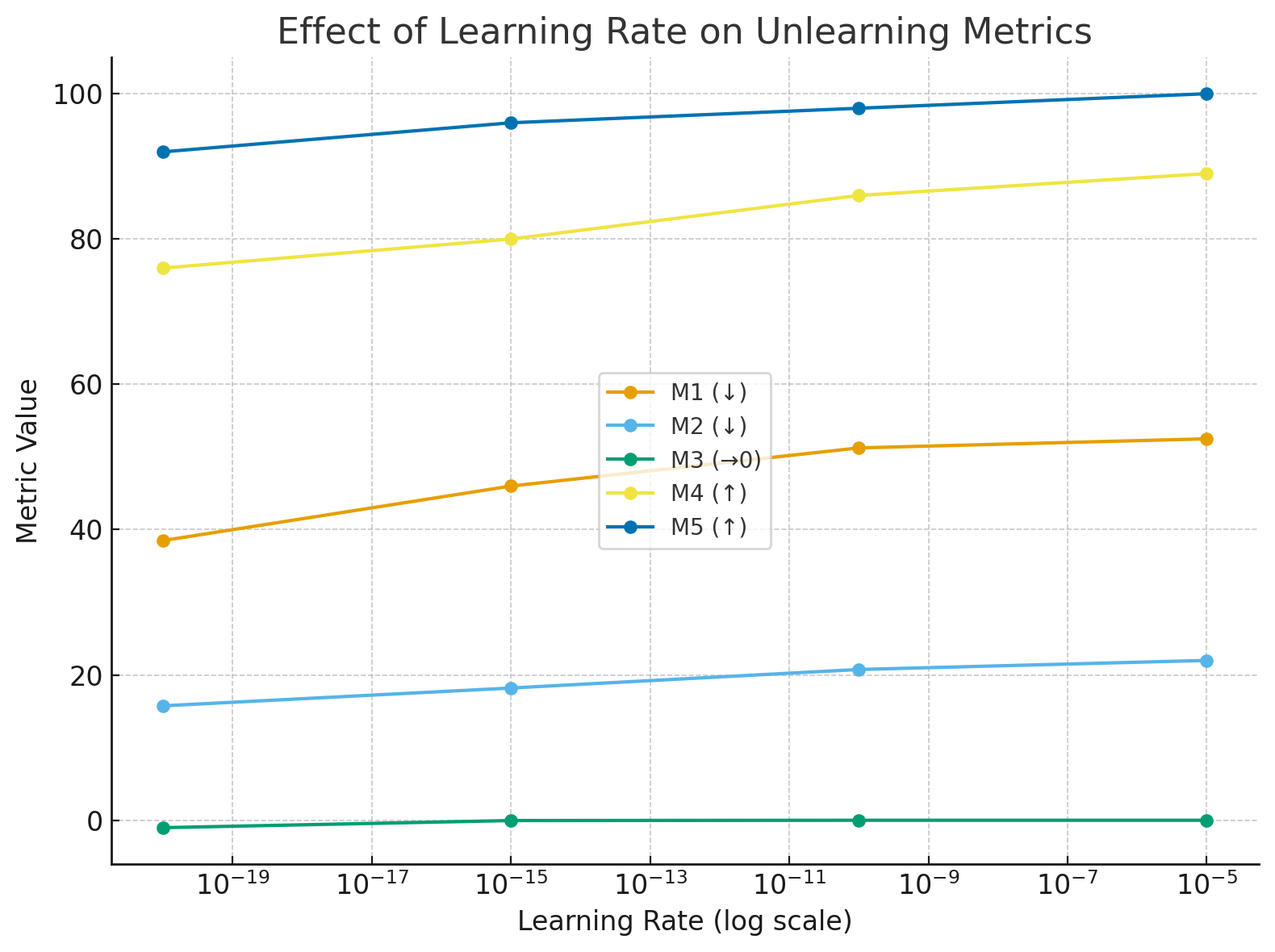}
   \caption{ Effect of learning rate change (1e-5 to 1e-20) on unlearning performance across M1–M5 metrics.}
   \label{fig:learn-rate}
\end{figure}

Figure~\ref{fig:learn-rate}, shows how the learning rate governs the balance between forgetting strength, retention quality, and stability within the RAZOR framework. Larger learning rates improve forgetting (lower M1/M2) while still maintaining strong utility (higher M4/M5), whereas very small learning rates weaken overall unlearning performance. This sensitivity provides a useful knob for prioritizing specific metrics depending on the requirements of a given unlearning scenario.

\section{Additional Evaluations}
\label{app:usecase}

\begin{figure}[t!]
   \includegraphics[width=0.99\linewidth]{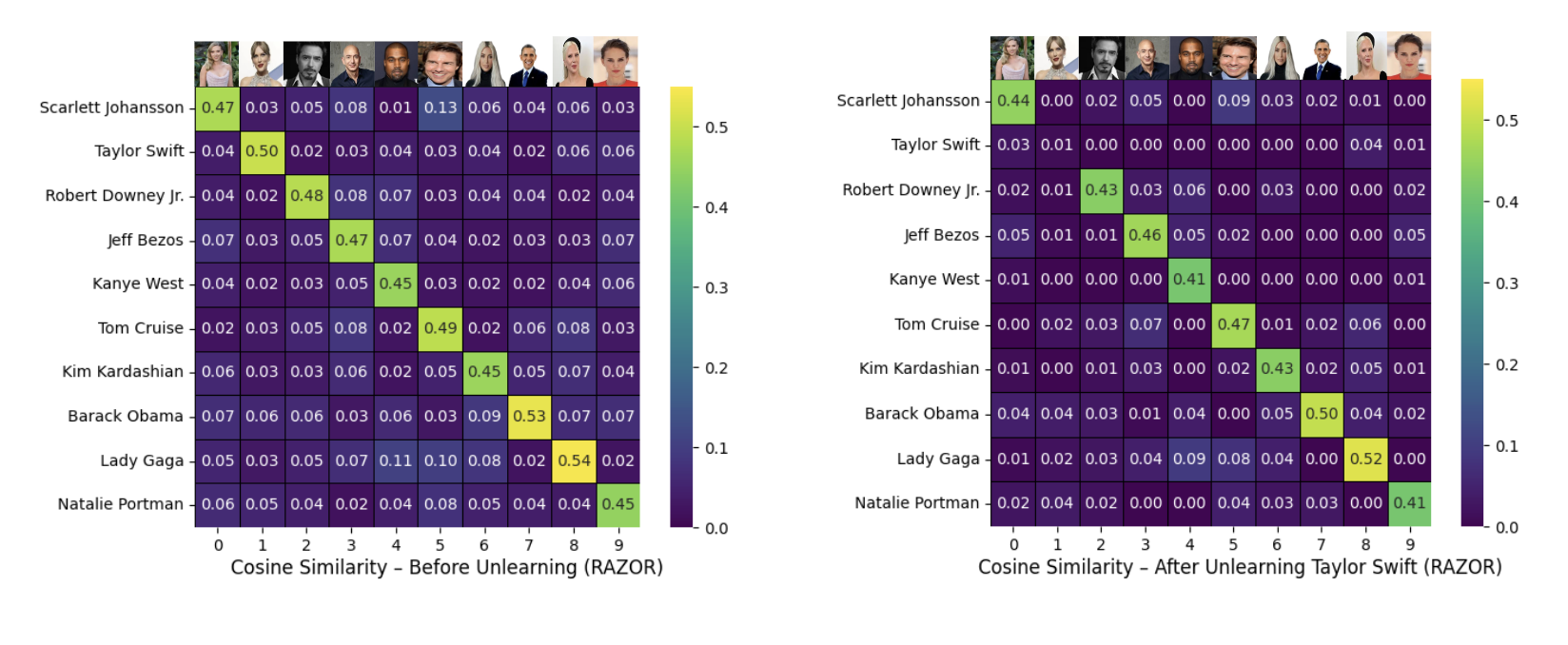}
   \caption{Cosine-similarity matrices before and after unlearning Taylor Swift using RAZOR on CLIP.}
   \label{fig:clip-cosign}
\end{figure}

Figure \ref{fig:clip-cosign} visualizes how RAZOR affects CLIP’s embedding space before and after unlearning the identity “Taylor Swift.” Prior to unlearning, Taylor Swift exhibits high self-similarity, indicating a strong and well-encoded identity representation in the model. After applying RAZOR, the similarity values for Taylor Swift collapse toward near-zero, demonstrating effective removal of her identity from the embedding space. Importantly, the similarity patterns for all other identities remain largely unchanged, confirming that RAZOR’s edits are highly localized and do not distort unrelated representations. This illustrates the method’s ability to selectively forget a target identity while preserving the semantic structure of the remaining identities.

\end{document}